\documentclass{bmvc2k}

\newcommand\blfootnote[1]{%
  \begingroup
  \renewcommand\thefootnote{}\footnote{#1}%
  \addtocounter{footnote}{-1}%
  \endgroup
}

\title{Temporally Compressed 3D Gaussian Splatting for Dynamic Scenes}

\addauthor{Saqib Javed}{saqib.javed@epfl.ch}{1$\dagger$} 
\addauthor{Ahmad Jarrar Khan}{ahmad.khan@epfl.ch}{1$\dagger$}
\addauthor{Corentin Dumery}{corentin.dumery@epfl.ch}{1}
\addauthor{Chen Zhao}{chen.zhao@epfl.ch}{1}
\addauthor{Mathieu Salzmann}{mathieu.salzmann@epfl.ch}{1, 2}

\addinstitution{
 CVLab, EPFL\\
 \small{Lausanne, Switzerland}
}
\addinstitution{
 Swiss Data Science Center\\
 \small{Lausanne, Switzerland}
}

\runninghead{Javed, Khan, Dumery, Zhao, Salzmann}{}

\newif\ifdraft
\drafttrue

\definecolor{orange}{rgb}{1,0.5,0}
\definecolor{green0}{rgb}{0.1,0.7,0.1}

\ifdraft
\newcommand{\MS}[1]{{\color{red}{\bf MS: #1}}}

\newcommand{\CD}[1]{{\color{blue}{\bf CD: #1}}}

\newcommand{\SJ}[1]{{\color{green0}{\bf SJ: #1}}}

\newcommand{\AK}[1]{{\color{violet}{\bf AK: #1}}}

\else
\newcommand{\MS}[1]{{\color{red}{}}}

\newcommand{\CD}[1]{{\color{blue}{}}}

\newcommand{\SJ}[1]{{\color{green0}{}}}

\newcommand{\AK}[1]{{\color{violet}{}}}

\fi

\usepackage{cleveref}
\usepackage{amsfonts}
\usepackage{dsfont}
\usepackage{algorithm}
\usepackage{algpseudocode}
\usepackage{booktabs}
\usepackage{colortbl}
\usepackage{graphicx}
\usepackage{wrapfig}
\usepackage{caption}
\usepackage{multirow}
\usepackage{subcaption}
\usepackage{colortbl}  
\usepackage{xcolor}
\usepackage{caption}

\begin{document}

\maketitle

\maketitle
\begin{figure}[h]
\vspace{-0.95cm}
    \centering
    \begin{tabular}{c@{\hspace{-0.5em}}c@{\hspace{-0.5em}}c@{\hspace{-0.5em}}c@{\hspace{-0.5em}}c}
 
        \vspace{-1.0em}
        \includegraphics[width=0.2\textwidth]{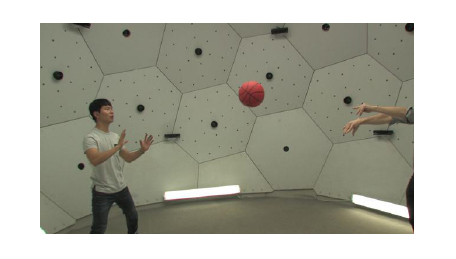} & 
        \includegraphics[width=0.2\textwidth]{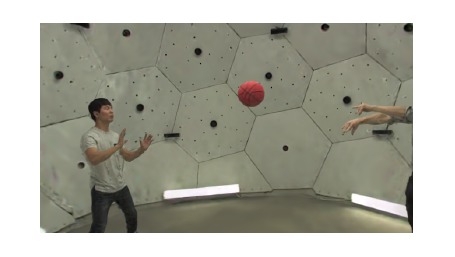} & 
        \includegraphics[width=0.2\textwidth]{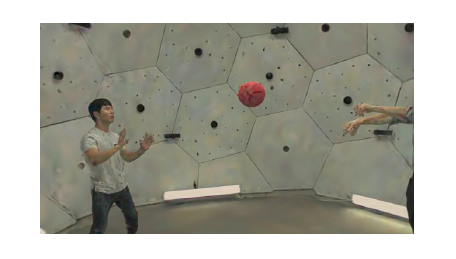} & 
        \includegraphics[width=0.2\textwidth]{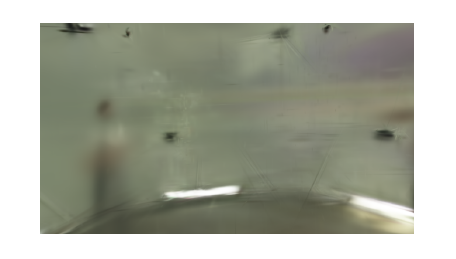} & 
        \includegraphics[width=0.2\textwidth]{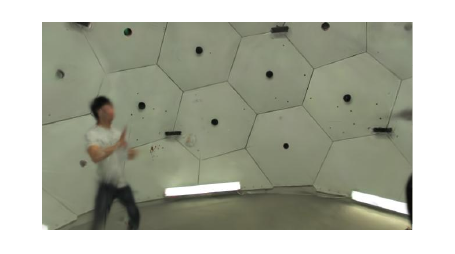} \\ 

        \vspace{-1.0em}
        \includegraphics[width=0.2\textwidth]{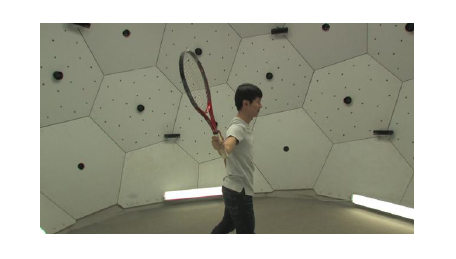} & 
        \includegraphics[width=0.2\textwidth]{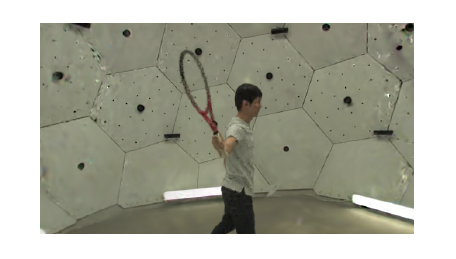} & 
        \includegraphics[width=0.2\textwidth]{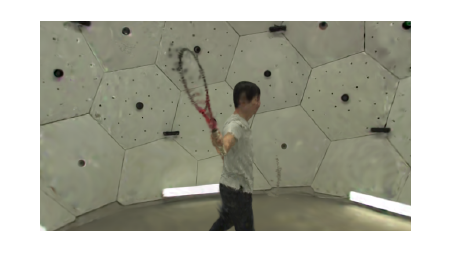} & 
        \includegraphics[width=0.2\textwidth]{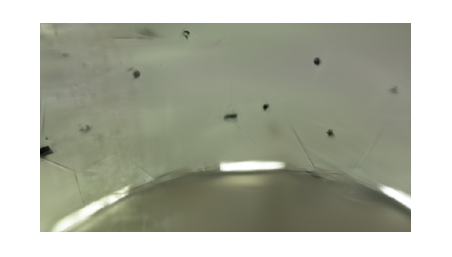} & 
        \includegraphics[width=0.2\textwidth]{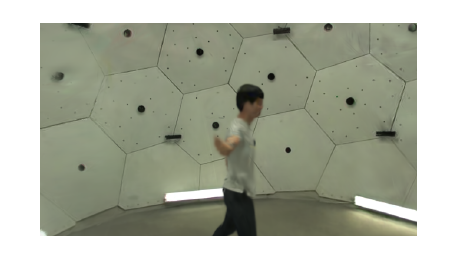}\\ 
        \vspace{-0.5em} \\
        \footnotesize{Ground Truth}  & \small{D-3DGS~\cite{luiten2023dynamic}}  & \footnotesize{TC3DGS} \footnotesize{(Ours)}  &  \footnotesize{STG~\cite{stg}}  & \footnotesize{4D Gaussian~\cite{4dgs}} \\
    \end{tabular}
    \vspace{0.2cm}
\captionsetup{width=0.9\linewidth}
\caption{\footnotesize{\textbf{Comparative Evaluation on Panoptic Dataset~\cite{Joo_2019}} Recent lightweight state-of-the-art methods (STG~\cite{stg}, and 4D-Gaussian~\cite{4dgs}) struggles to reconstruct scenes in complex environments accurately. In contrast, our compression strategy effectively captures and maintains high fidelity.}}
    \label{fig:Comparison}
\end{figure}

\vspace{-0.9cm}

\begin{abstract}
\vspace{-0.2cm}
Recent advancements in high-fidelity dynamic scene reconstruction have leveraged dynamic 3D Gaussians and 4D Gaussian Splatting for realistic scene representation. However, to make these methods viable for real-time applications such as AR/VR, gaming, and rendering on low-power devices, substantial reductions in memory usage and improvements in rendering efficiency are required. While many state-of-the-art methods prioritize lightweight implementations, they struggle in handling {scenes with complex motions or long sequences}. In this work, we introduce Temporally Compressed 3D Gaussian Splatting (TC3DGS), a novel technique designed specifically to effectively compress dynamic 3D Gaussian representations. TC3DGS selectively prunes Gaussians based on their temporal relevance and employs gradient-aware mixed-precision quantization to dynamically compress Gaussian parameters.
In addition, TC3DGS exploits an adapted version of the Ramer-Douglas-Peucker algorithm to further reduce storage by interpolating Gaussian trajectories across frames.
Our experiments on multiple datasets demonstrate that TC3DGS achieves up to 67$\times$ compression with minimal or no degradation in visual quality. More results and videos are provided in the supplementary.\blfootnote{$\dagger$ Equal contribution.}

Project Page: \textcolor{magenta}{\url{https://ahmad-jarrar.github.io/tc-3dgs/}}

\end{abstract}   

\vspace{-0.55cm}
\section{Introduction}
\label{sec:intro}
\vspace{-0.1cm}
Dynamic scene reconstruction is essential for applications in virtual and augmented reality, gaming, and robotics, where a real-time and accurate representation of moving objects and their environment is the key to immersive experiences. Recent progress such as Neural Radiance Fields (NeRF)~\cite{mildenhall2021nerf} has enabled high-fidelity scene generation, although at the cost of a slow rendering speed. To address this limitation, 3D Gaussian Splatting (3DGS)~\cite{kerbl20233d} leverages sparse Gaussian splats for efficient scene rendering, particularly for static scenes.

Since the advent of 3DGS, various methods have been proposed to extend dynamic scene modeling, either by evolving Gaussians over time to capture time-varying properties~\cite{4dgs,luiten2023dynamic,das2024neural,yang2024deformable,xu20234k4d} or by learning spatio-temporal Gaussians for more flexible scene representation~\cite{das2024neural, yang2024deformable,duan20244d,yang2023real,Jiang_2024_CVPR}. Our experiments with such frameworks, illustrated in Figure~\ref{fig:Comparison} and in supplementary videos have highlighted that many state-of-the-art methods struggle to effectively adjust Gaussian parameters in dynamic scenes with rapid and complex motions, such as those of~\cite{Joo_2019}. The only exception is Dynamic 3D Gaussians~\cite{luiten2023dynamic}, which mitigates this issue by enforcing consistency across frames. However, this comes with a dramatic increase in storage and rendering costs, making this method impractical for real-world applications, such as deployment on VR/AR headsets.

To address these challenges, we propose \textbf{Temporally Compressed 3D Gaussian Splatting (TC3DGS)}, a novel approach designed to efficiently compress dynamic 3D Gaussian representations for high-quality, real-time scene rendering. Unlike traditional methods, TC3DGS reduces both the number and the memory footprint of the Gaussians by selectively pruning splats based on temporal importance and learning a parameter-specific bit-precision. This selective compression allows us to maintain scene fidelity while significantly reducing storage and computational requirements, making TC3DGS well-suited for dynamic environments with complex motions.

Our approach begins with a pruning and masking strategy designed to eliminate redundant Gaussians. Although some works \cite{zhang2024lp,fan2023lightgaussian,lee2024compact} have proposed pruning methods for 3DGS, they do not attempt to model dynamic scenes, and thus take no advantage of temporal compression. Unfortunately, adapting these pruning strategies to dynamic scenes is not straightforward, as different Gaussians may need to be pruned across different frames.
Here, we introduce a method that explicitly
integrates this temporal aspect into the training objective, allowing us to prune dynamic Gaussian splats more effectively. 
To further optimize memory usage in our scene representation, we also compress the storage size of the remaining Gaussians by developing a gradient-aware mixed-precision quantization method that adjusts the bit precision of each Gaussian parameter based on its \textit{sensitivity}. We use gradient magnitudes to determine parameters with high sensitivity, and allocate them a higher precision, while those with lower impact on the scene are quantized with fewer bits. This allows our method to achieve a good balance between compression and reconstruction accuracy. Finally, to further enhance the efficiency of our dynamic representation, we introduce a keypoint extraction algorithm as a post-processing step, which leads to an overall compression rate of up to 67$\times$ while preserving rendering quality. 

Our experiments on benchmark datasets demonstrate the benefits of our method on diverse scenarios. Furthermore, we perform an ablation study to demonstrate the contribution of each key component in our pipeline. In a nutshell, our major contributions are as follows:
\begin{itemize}
    \item We identify and analyze the failure of existing methods in handling scenes with rapid and complex motion, highlighting these shortcomings through qualitative and quantitative evaluations.
  \item We introduce the first method to prune dynamic Gaussian splats, going beyond previous pruning techniques, designed for static scenes, by incorporating temporal relevance into the pruning process.
  \item We develop a sensitivity-driven, gradient-based mixed-precision quantization method that dynamically assigns bit precision to parameters based on their impact on reconstruction accuracy, optimizing memory usage.
  \item We propose a keypoint extraction post-processing algorithm to further reduce storage requirements by simplifying the Gaussian parameter trajectories, retaining only essential data points for compact scene representations.
\end{itemize}


\section{Related Work}
\label{sec:related_work}
\vspace{-0.2cm}
\paragraph{Dynamic 3D reconstruction.}
Recent advancements in 3D reconstruction based on Neural Radiance Fields (NeRFs) \cite{mildenhall2021nerf} and 3D Gaussian Splatting (3DGS) \cite{kerbl20233d} have achieved remarkable levels of visual fidelity and accuracy. 
These methods have subsequently been extended to 4D representations \cite{pumarola2021d,liu2023robust, TiNeuVox, kulhanek2024wildgaussians, Li2024ST}, enabling dynamic scene reconstruction. Methods to decompose a 4D scene into multiple 2D planes to learn a more compact representation are also explored by various methods~\cite{kplanes_2023, Cao2023HexPlane, shao2023tensor4d, attal2023hyperreel}.
In the case of 3DGS, where Gaussians are explicitly stored and rendered, different approaches have emerged to model their time dependence. One prominent line of work \cite{4dgs,luiten2023dynamic,das2024neural,yang2024deformable, huang2024scgs, kratimenos2024dynmf} in this area optimizes a canonical set of Gaussians from the initial frame, and combines it with a deformation motion field allowing temporal variations of the Gaussian parameters. However, these methods are limited to short videos, as they cannot add Gaussians after the initial frame.Recent works \cite{sun20243dgstream, hicom2024} have explored online causal training of 3DGS models for streaming applications and to handle long scenes. However, these methods scale with the number of frames, like Dynamic 3DGS \cite{luiten2023dynamic}.

Another class of methods \cite{yang2023real,duan20244d,katsumata2025compact} directly models temporal Gaussians that can be present in a subset of frames, enabling certain elements to appear in selected time ranges and thus increasing the expressivity of their reconstruction. However, a major limitation across these methods is that both training and inference times for novel view synthesis scale with the number of Gaussians, the length of the sequence, and the complexity of their parameters, presenting a key bottleneck in enhancing reconstruction quality.

\vspace{-0.4cm}
\paragraph{Compressed 3D radiance fields.}
An important research direction has thus emerged in developing more compact representations of radiance fields. For NeRFs, compact grid structures \cite{muller2022instant,chen2022tensorf,fridovich2022plenoxels,franke2024trips, scaffoldgs} have already proven effective in reducing network sizes and enhancing accuracy.

With 3DGS, recent works have either concentrated on optimizing the representation of the Gaussian parameters \cite{fan2023lightgaussian, chen2024fast}, or on identifying low-importance Gaussians and pruning them entirely \cite{zhang2024lp,fan2023lightgaussian,lee2024compact,girish2024eaglesefficientaccelerated3d}. Unorthodox techniques like representing the Gaussian parameters as 2D grids and applying image compression techniques~\cite{morgenstern2024self_grid, navaneet2024compgs} have also been studied. Better initialization and weighted sampling based pruning~\cite{Fang2024MiniSplattingRS} has also shown promising results. Additionally, the use of traditional compression techniques such as vector quantization \cite{lee2024compact,navaneet2024compgs} or entropy models \cite{chen2025hac} have shown some potential for the compression of static scenes, but scaling them to dynamic scenes with possibly hundreds or thousands of frames remains a challenge.

Indeed, dynamic scenes require an even larger set of parameters to accurately capture motion, temporal variations, and complex interactions within the scene. 
The temporal relevance of each Gaussian changes dynamically, and traditional pruning strategies designed for static scenes are insufficient, as they lack adaptability to these fluctuations. To the best of our knowledge, we are the first to propose a compression framework specifically designed for dynamic 3D Gaussians. Our approach combines temporal relevance-based pruning, gradient-based mixed-precision quantization, and trajectory simplification to address the unique requirements of dynamic scenes.

\vspace{-0.5cm}
\section{On the Challenges of Fast Motion}
\vspace{-0.2cm}
One of the most widely used datasets for benchmarking dynamic novel view synthesis, the Neural 3D Video dataset~\cite{li2022neural3dvideosynthesis}, consists exclusively of indoor scenes with objects moving within their local vicinity. Existing methods thus perform well, as object motion remains limited. However, when we applied these methods to the Panoptic Sports dataset~\cite{Joo_2019}, their performance deteriorated significantly due to the presence of fast moving objects that travel longer distances. Below, we analyze the underlying reasons why these methods struggle on such complex datasets.

Existing Gaussian Splatting methods for dynamic scenes often incorporate motion priors to regulate the movement of the Gaussian points. 
These priors determine the type of motion for which the method is more suitable.
The key distinction between different approaches is how they integrate these motion priors. 

Specifically, deformation field-based methods, such as 4D-Gaussians~\cite{4dgs} and E-D3DGS~\cite{bae2024ed3dgs}, implicitly encode motion priors within the structure of neural networks, yielding flexible but less interpretable motion representations. These methods fails in the presence of large sudden movements due to the \textit{spectral bias} of neural networks towards simpler functions \cite{rahaman2019spectral}.

As an alternative, STG~\cite{stg} models point trajectories as polynomial curves. However, this approach restricts the representation to 3\textsuperscript{rd} degree polynomials, which limits the types of motion it can represent. 
While higher-degree polynomials enable the representation of more complex motion, they also introduce significant storage overhead. 

{Deformable 3DGS~\cite{yang2024deformable} impose explicit constraints, e.g., local rigidity, to regulate motion in a more interpretable manner. However, the use of a neural network in Deformable 3DGS to model the deformation field inherits the limitations of 4D-Gaussians and E-D3DGS discussed earlier.}

Among the methods discussed above only Dynamic 3DGS~\cite{luiten2023dynamic}{, deformable field and explicit constraints based model,} is trained causally, i.e., learning Gaussian deformations frame by frame, instead of optimizing the Gaussians across all time frames simultaneously, as done by the other methods. This causal training strategy requires storing and optimizing point positions in every frame, offering greater flexibility in motion modeling despite the local rigidity constraint, as the Gaussians are optimized independently for each frame. However, this comes at a high storage and rendering cost. Therefore, in~\cref{sec:compress}, we propose a method to significantly compress Dynamic 3DGS while preserving its ability to capture complex motions.

In Figure~\ref{fig:Comparison}, we provide a visual comparison illustrating how existing methods struggle with scenes featuring rapid and complex motion. In addition, we provide videos in the supplementary material for a more extensive evaluation.

\vspace{-0.0cm}
\section{Compression}\label{sec:compress}
\begin{figure*}
\vspace{-0.1cm}
    \centering
    \includegraphics[width=\linewidth]{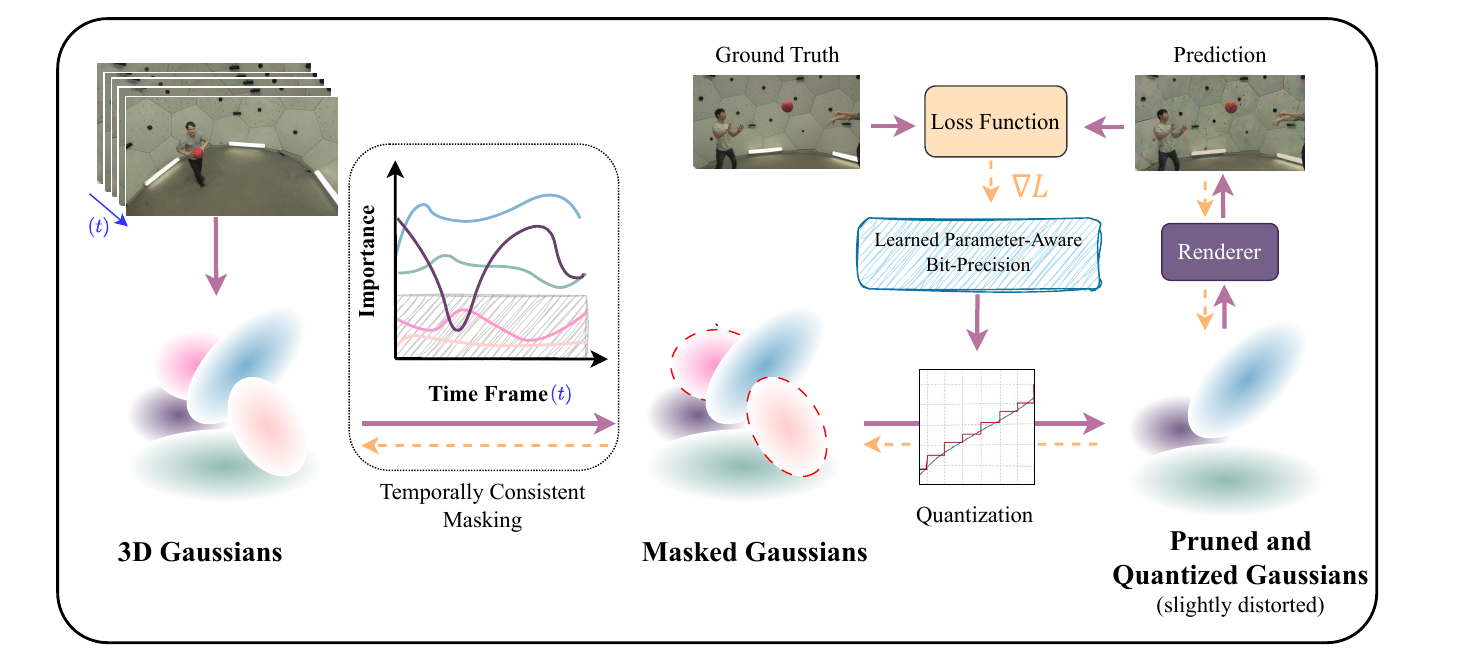}
    \vspace{-0.3cm}
    \caption{\small{\textbf{Overview of our Temporally Compressed 3D Gaussian Splatting for Dynamic Scenese (TC3DGS) method.} Our approach involves a temporally consistent masking strategy to select relevant 3D Gaussians across frames. The masked Gaussians are then pruned and quantized using a gradient-based, parameter-aware bit-precision quantization scheme.}\\
    }
    \label{fig:Main}
    \vspace{-1cm}
\end{figure*}

\vspace{-0.2cm}
In this section,  
we introduce our TC3DGS approach, encompassing
novel masking and pruning strategies designed to eliminate redundant Gaussians, a sensitivity-based mixed precision technique for efficient parameter compression,
and a post-training compression strategy to minimize storage overhead. The  overview of our method can be visualized in Figure~\ref{fig:Main}.

\vspace{-0.3cm}
\subsection{TC3DGS}
Dynamic 3DGS \cite{luiten2023dynamic} is particularly promising because it models the dynamic scene as movements of Gaussians under kinematic constraints w.r.t. the previous timestep. By optimizing the position and rotation of the Gaussians instead of learning deformation functions, it removes the limitation on possible deformations due to the characteristics of the modeling function. However, by learning position and rotation at each timestep separately, the number of parameters increases linearly, resulting in large storage sizes, and thus limiting its applicability to high-fidelity and long-range scene modeling.

\begin{wrapfigure}[10]{t}{0.5\linewidth}
    \centering
    \vspace{-1cm}
    \scalebox{0.97}{
    \includegraphics[width=\linewidth]{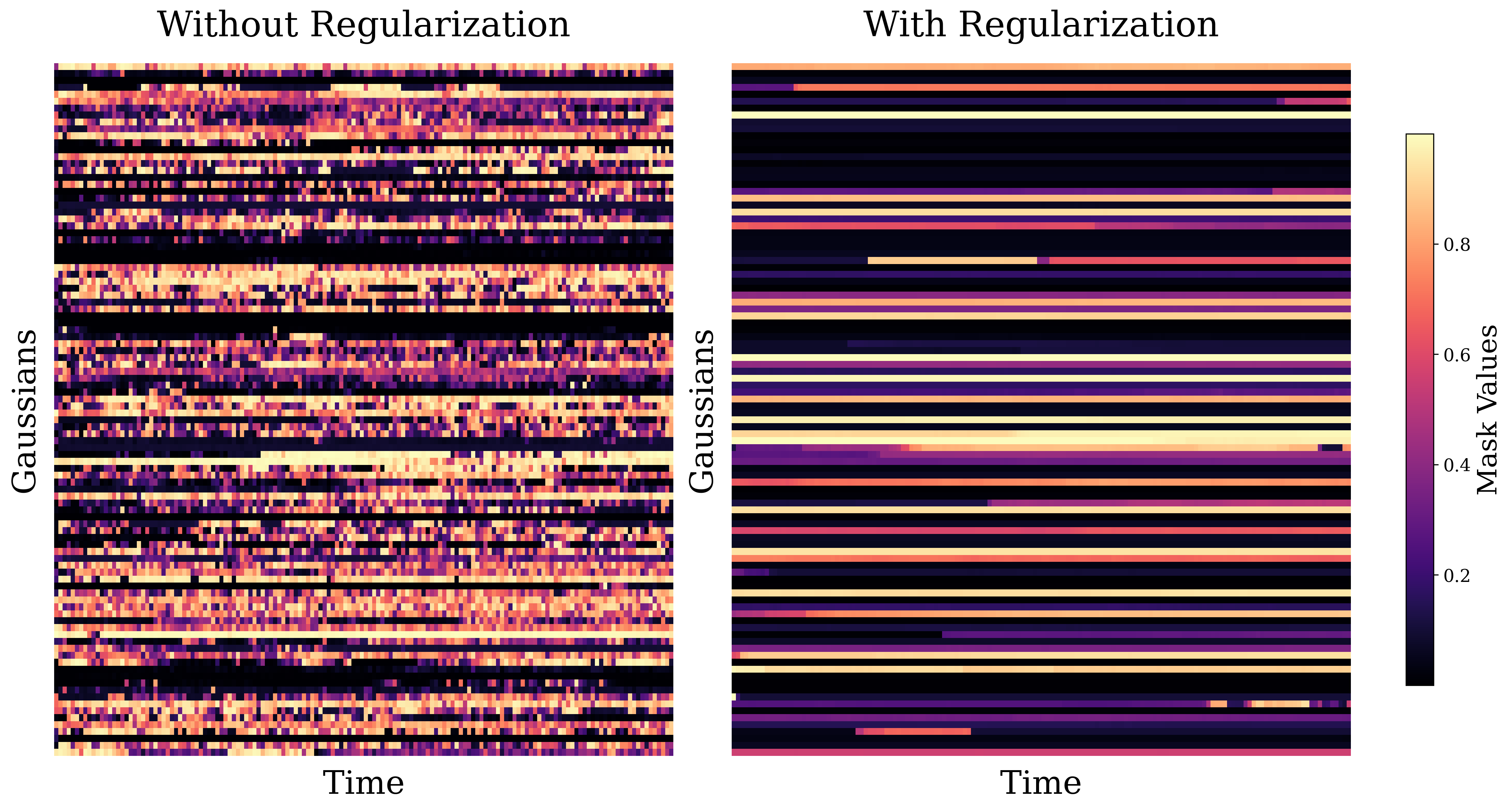}}
    \vspace{0.15cm}
    \caption{\small{\textbf{Mask Consistency.} (Left) Mask values of Gaussians trained independently for each time frame. (Right) Mask values trained with our loss.}}
    \label{fig:mask_consistency}
\end{wrapfigure}
\subsubsection{Gaussian Masking and Pruning}\label{sec:pruning}
\vspace{-0.15cm}
Pruning techniques aimed at identifying and removing low-importance Gaussians have been successfully applied to static 3D Gaussian splatting \cite{lee2024compact, Niedermayr_2024_CVPR, girish2024eaglesefficientaccelerated3d, fan2023lightgaussian}. However, for dynamic scenes, a temporally consistent importance measure is required to ensure that the pruned Gaussians remain insignificant throughout the scene duration.

Various approaches to computing the importance of each Gaussian in static scenes using training images and camera positions have bee proposed~\cite{Niedermayr_2024_CVPR, fan2023lightgaussian}. These methods focus on the contribution of each Gaussian to the training views. In dynamic scenes, Gaussians are not stationary, so their contributions vary over time. To prune Gaussians effectively from dynamic scenes, it is essential to maintain the contributions of high-importance Gaussians consistently high, while suppressing low-importance ones, thus enabling more effective pruning.

Compact-3DGS~\cite{lee2024compact} introduces a masking approach based on Gaussian volume and transparency. Gaussians with low opacity, minimal volume, or both are masked out because they have negligible impact on the rendered images. A mask parameter $m \in \mathbb{R}^N$ is learned to produce binary masks $M \in \{0, 1\}^N$ using a straight-through estimator. This binary mask is then applied to the Gaussians by scaling their opacities and sizes. This is expressed as
\vspace{-0.1cm}
\begin{align}
    M_n &= \operatorname*{sg}(\mathds{1}[\sigma(m_n) > \epsilon] - \sigma(m_n)) + \sigma(m_n), \label{eq:Mn} \\
    \hat{s}_n &= M_n s_n,\quad \hat{o}_n = M_n o_n,
\end{align}

\noindent where $n$ represents the Gaussian index, $\epsilon$ denotes the masking threshold, $\operatorname*{sg}(\cdot)$ is the stop-gradient operator, {$s_n \in \mathbb{R}^3$ represents scale of the gaussian, $o_n \in \mathbb{R}$ represents opacity of the gaussian,} and $\mathds{1}[\cdot]$ and $\sigma(\cdot)$ correspond to the indicator and sigmoid functions.

In~\cite{lee2024compact}, it is explained that this mask learns to remove Gaussians with low opacity and/or small volume. However, as shown in Figure~\ref{fig:mask_consistency}, this explanation does not hold in dynamic scenes. The value of $m_n$ can change significantly across time frames, even with fixed opacity $o$ and scale $s$ after the initial time frame.

In dynamic scenes, $m_n$ depends on the 2D projected area of the Gaussians in the training views and their transmittance, where transmittance $T_i$ for the $i^{th}$ Gaussian along a camera ray is defined as the Gaussian's contribution to blending, \textit{i.e.},
\vspace{-0.2cm}
\begin{equation}
T_i = \sigma(o_i) \prod_{j=1}^{i-1}(1 - \sigma(o_j)).
\end{equation}
\vspace{-0.1cm}
As a Gaussian moves relative to the others, changes in $T_i$ are reflected in $m_n$. Similarly, when a Gaussian moves toward or away from the training cameras, its projected area in the views changes, affecting $m_n$ accordingly.

Since the values in $m$ strictly need to be high due to their role in rendering through \cref{eq:Mn} and the photometric loss, we introduce an additional regularization to incentivize lower values. We thus regularize $m$ by minimizing
\vspace{-0.2cm}
\begin{equation}
    \mathcal{L}_{mask} = \sum_{n=1}^{N} \sigma(m_n).
    \label{eq:mask_loss}
\end{equation}
This regularization loss 
penalizes unnecessarily high values, reducing $m$ to the minimum required to produce satisfying renderings. The flexibility of this learned mask is one of its key advantages, as it can be optimized to exhibit desired properties via the use of additional constraints.

A key motivation for ensuring consistency of {$m_{n}$} across frames is to capture the global importance of the Gaussians during pruning. To achieve this, we {replace $m_n$ with a time dependent $m_{n,t}$ defined for every timestamp separately and use it instead of $m_n$ in the rendering. We introduce} a consistency loss function that encourages $m_{n,t}$ to remain close to $m_{n,t-1}$.
Specifically, we define this mask consistency loss function  as
\vspace{-0.2cm}
\begin{equation}
\mathcal{L}_{mc} = \sum_{n=1}^{N} |m_{n,t} - sg(m_{n,t-1})|.
\label{eq:mask_cons_loss}
\end{equation}
It ensures that the masks exhibit stability across the frames, as shown in Figure \ref{fig:mask_consistency} (Right). By maintaining consistency, our approach prevents sudden fluctuations in Gaussian importance, which can degrade rendering quality and lead to suboptimal pruning results.

After optimizing across all timestamps, we perform Gaussian pruning based on the average value of $m_n$. Specifically, a Gaussian is pruned if its average value across all timestamps satisfies
\vspace{-0.2cm}
\begin{equation}
\frac{1}{T} \sum_{t=1}^{T} \sigma(m_{n,t}) < \epsilon.
\end{equation}
\vspace{-0.4cm}
\subsubsection{Gradient-Aware Mix-Precision Quantization}\label{sec:quant}
\vspace{-0.2cm}

The influence of the Gaussian parameters on reconstruction quality is highly variable; a small adjustment in one parameter can significantly alter the rendered image, whereas similar adjustments in other parameters or in the same parameter of a different Gaussian may have minimal impact. Our approach uses gradient-based sensitivities to dynamically assign bit precision to each parameter, based on its influence on reconstruction accuracy. By leveraging this adaptive, in-optimization quantization, each parameter adjusts its quantization scale~\cite{Esser2020LEARNED} in real-time, preserving detail in the reconstructed scene.

We first calculate the mean sensitivity for each parameter based on the gradients~\cite{Zhang_2024,Niedermayr_2024_CVPR}, reflecting each parameter's contribution on the image reconstruction performance. We introduce a sensitivity coefficient, 
formulated as
\begin{align}\label{eq:sensitivity}
\vspace{-0.3cm}
    S(\theta) &= \frac{1}{\sum_{k=1}^K N_k} \sum_{k=1}^K \big{|} \frac{\partial Q_k}{\partial \theta} \big|.
\end{align}

\noindent where \( K \) is the number of training images used for reconstruction, \( N_k \) is the number of pixels in the $k^{th}$ image, and \( Q_k \) denotes the cumulative pixel intensity across the RGB channels in image \( k \).

The coefficient \( S(\theta) \) quantifies the average change in pixel intensity from training views when parameter \( \theta \) is changed by a small amount.

By using this impact-based metric, we can effectively rank the parameters by their importance on image fidelity. {We then normalize each sensitivity coefficient by scaling it based on the minimum and maximum co-efficient across all parameters. This standardization ensures that all the coefficients fall within a consistent range. Therefore, we allocate higher bit precision to more sensitive parameters and lower bit precision to less sensitive ones, optimizing the balance between computational efficiency and model accuracy.}

After running our scene reconstruction process for a specified number of iterations, we apply mixed-precision quantization to all Gaussian parameters, excluding the position parameter, to achieve low-bit precision for the other parameters. Instead of relying on traditional vector quantization (VQ) or basic min-max quantization, we propose a parameter quantization technique with learnable scaling factors~\cite{Esser2020LEARNED}, integrating it directly into the optimization process rather than treating it as a post-optimization fine-tuning step.
Consequently, many parameters can be effectively quantized to 4-bit precision, reducing memory and computational load without compromising reconstruction quality.

\vspace{0.5em}
\textbf{Training.} 
We train the Gaussian parameters to model the scene one time frame at a time. Following \cite{luiten2023dynamic}, we use physically-based priors to regularize the Gaussians. The optimization objective is defined as
\vspace{-0.2cm}
\begin{align}
    \mathcal{L} = \mathcal{L}_{original} 
    + \lambda_{mask}\mathcal{L}_{mask} 
    + \lambda_{mc}\mathcal{L}_{mc} ,
\end{align}

\noindent where $\mathcal{L}_{original}$ is the same loss function as in \cite{luiten2023dynamic}, while $\mathcal{L}_{mask}$ and $\mathcal{L}_{mc}$ are defined in Eq. \ref{eq:mask_loss} and \ref{eq:mask_cons_loss}, respectively.

\vspace{1.4cm}
\begin{wrapfigure}[11]{t}{0.5\textwidth}
    \vspace{-0.8cm} 
    \begin{minipage}{\linewidth}
    \footnotesize{
        \captionsetup{type=figure}
        \caption*{\textbf{Algorithm 1:} Keypoint Selection}
        \vspace{0.2cm}
        \textbf{Input:} values, max\_keypoints ($max_{kp}$), tolerance ($\tau$) \\
        \vspace{-0.2cm}
        \begin{algorithmic}[1]
            \State Initialize keypoints at first and last positions
            \For{$i = 1$ to $max_{kp} - 2$}
                \State Compute interpolated values based on keypoints
                \State Compute error for every value
                \State Compute error (\texttt{mse}) for entire sequence
                \If{\texttt{mse} $\leq \tau$ \textbf{or} $|\texttt{keypoints}| \geq max_{kp}$}
                    \State \textbf{break}
                \EndIf
                \State Select the value with the highest error and add it to keypoints
            \EndFor
        \end{algorithmic}
        }
    \end{minipage}
    \vspace{1pt}
\end{wrapfigure}

\vspace{0.2cm}
\vspace{0.3cm}
\subsubsection{Keypoint Interpolation}\label{sec:keypoint}
\vspace{-0.15cm}

Dynamic 3DGS~\cite{luiten2023dynamic} opts for inefficiently storing all time-dependent parameters, such as Gaussian means, rotations, and colors, for all time frames. While this greatly increases the expressivity of this method compared to other works, it comes at a significant cost in memory.


\begin{wrapfigure}[15]{t}{0.5\textwidth}
    \centering
    \vspace{-0.4cm}
    \scalebox{0.97}{
    \includegraphics[width=\linewidth]{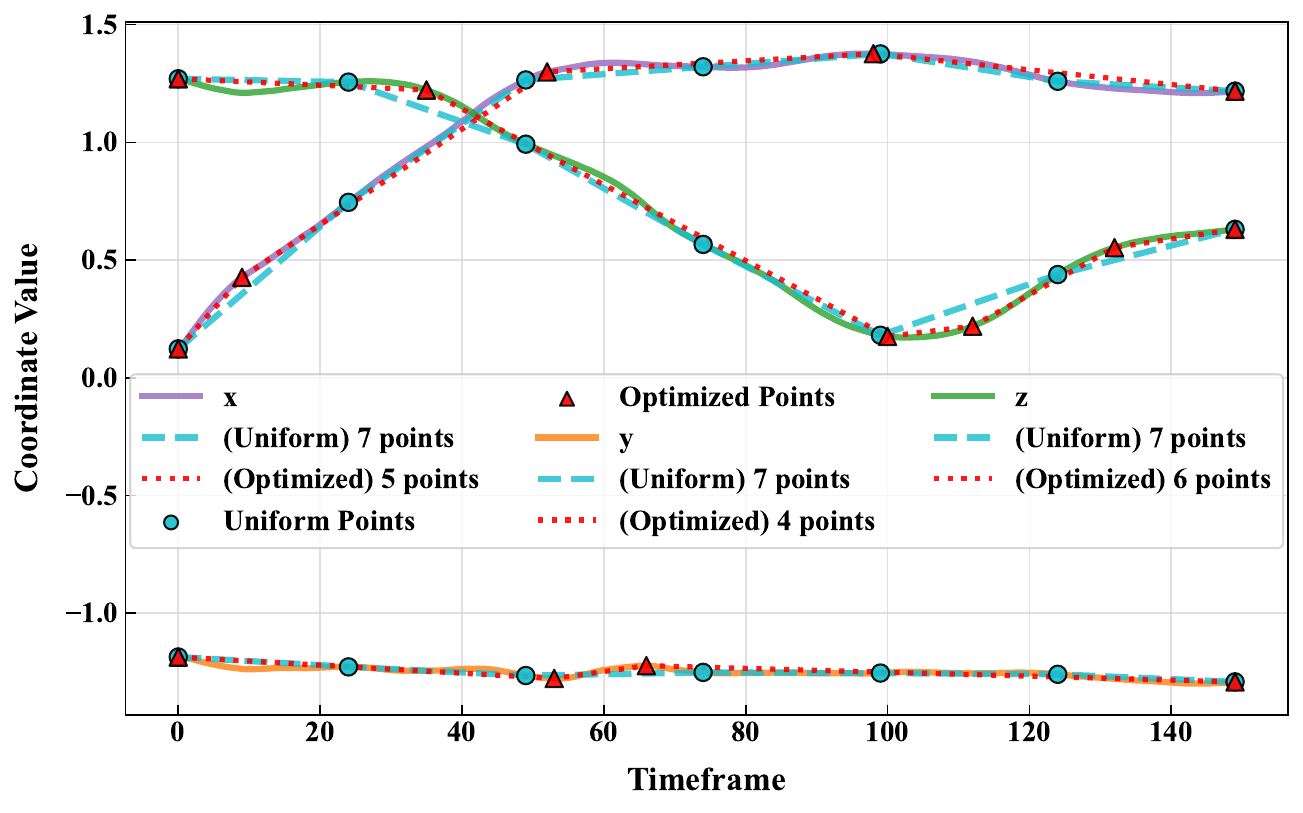}}
    \caption{\small{\textbf{Keypoint Interpolation.} In this example, we represent a position across 150 frames with only 5, 4 and 6 keypoints for $x$, $y$ and $z$, respectively, with only $0.038$ MSE. By comparison, uniformly sampling 7 keypoints increases storage and increases error to $0.089$ MSE.}}
    \vspace{-0.5cm}
    \label{fig:Keypoint}
\end{wrapfigure}

We take a different approach and observe that only a small subset of \textit{keypoints} are required to accurately reconstruct complex motions. Instead of storing each gaussian's position and rotation per frame, we identify keypoints where the trajectory of a gaussian changes significantly and interpolate linearly between them. However, the placement of these keypoints across time cannot be predetermined, as it depends on the individual Gaussians. For instance, background Gaussians require a single keypoint for the entire sequence, whereas moving objects will need substantially more keypoints. This motivates the development of our keypoint selection strategy, which we adapt from the Ramer–Douglas–Peucker (RDP) algorithm \cite{RDP}. It is applied as a post-processing step to further reduce storage requirements for the time-varying Gaussian parameters in dynamic scenes.

While the RDP algorithm selects keypoints from a sequence based on a local error tolerance, $\xi$, we propose a novel keypoint selection method, as outlined in Algorithm 1. 
It provides greater flexibility by allowing control via both an acceptable tolerance value, $\tau$, and a maximum number of keypoints $max_{kp}$. The parameter $\tau$ defines the maximum allowable Mean Squared Error (MSE) over the sequence. Unlike RDP, which is solely controlled by $\xi$, our method enables a hard maximum bound on the number of keypoints, allowing for more precise, fine-grained control. In Figure 4, we illustrate with an example how our method achieves lower MSE despite storing fewer keypoints.

Following this, we flatten and transpose the time-dependent parameters from $\mathbb{R}^{T \times N \times D}$ to $\mathbb{R}^{ND \times T}$, transforming the data into $ND$ sequences of length $T$. We then compute the keypoints for all Gaussians in parallel, forming sparse matrices. The sparsity of these matrices is controlled using the parameters $max_{kp}$ and $\tau$, and and we store these sparse matrices to minimize memory usage.

\vspace{-0.4cm}
\section{Experiments}
\vspace{-0.1cm}
 We evaluate the methods on diverse datasets covering different real-world scenarios. Results on additional datasets, along with dataset and implementation details, are provided in the supplementary material, which also includes visualizations. Moreover, ablation studies comparing our quantization strategy to uniform quantization are also included.

\begin{wrapfigure}[10]{t}{0.51\linewidth}
\vspace{-1.1cm}
    \centering
    \includegraphics[width=0.97\linewidth]{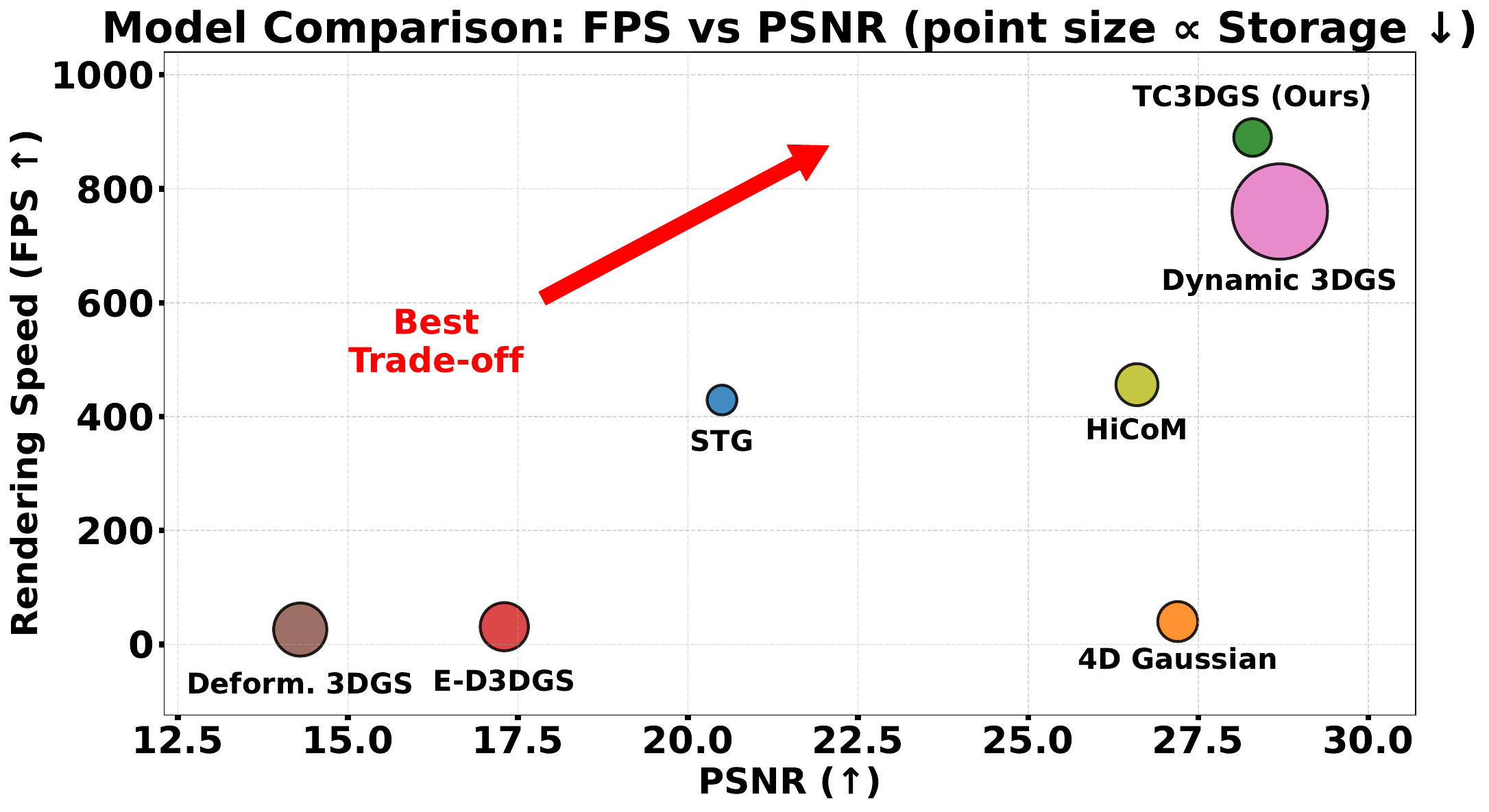}
    \vspace{0.2cm}
    \caption{\small{{Our method lies on the Pareto frontier, achieving competitive performance with significantly smaller model size.}\\}
    }
    \label{fig:Main2}
\end{wrapfigure}

\vspace{-0.4cm}
\subsection{Results}
\label{Results}

As shown in Table~\ref{tab:combined_results} and Figure~\ref{fig:Comparison}, our method achieves comparable results to Dynamic 3DGS~\cite{luiten2023dynamic}, while using, on average, 40 times less storage. Similarly, the results on the Neural 3D Video dataset in Table~\ref{tab:combined_results}
demonstrate competitive performance with a significantly smaller storage footprint and fastest rendering speed. 
These improvements are achieved without a substantial increase in training time compared to Dynamic 3DGS \cite{luiten2023dynamic}, as our method trains fewer Gaussians and just one additional parameter per Gaussian for masking.

\begin{table}[t]
    \centering
    \footnotesize
    \def\arraystretch{1.1}
    \scalebox{0.9}{
    \begin{tabular}{lcccccc}
        \toprule
        \multirow{2}{*}{Method} & \multicolumn{3}{c}{Panoptic Dataset} & \multicolumn{3}{c}{Neural 3D Video Dataset} \\
        \cmidrule(lr){2-4} \cmidrule(lr){5-7}
        & PSNR ↑ & FPS ↑ & Storage ↓ & PSNR ↑ & FPS ↑ & Storage ↓ \\
        \midrule
        STG & 20.5 & 429 & \textbf{19MB} & \textbf{32.04} & 273 & 175MB \\
        4D Gaussian & 27.2 & 40 & 62MB & 31.15 & 30 & 90MB \\
        E-D3DGS & 17.3 & 31 & 130MB & 31.20 & 69 & \underline{40MB} \\
        Deform. 3DGS & 14.3 & 26 & 192.6MB & 30.97 & 29 & \textbf{33MB} \\
        Dynamic 3DGS & \textbf{28.7} & \underline{760} & 1994MB & \underline{31.38} & \underline{460} & 2772MB \\
        HiCoM & 26.6 & 456 & 71MB & 31.17 & 247 & 270MB \\
        \rowcolor[gray]{0.92} TC3DGS (Ours) & \underline{28.3} & \textbf{890} & \underline{49MB} & 30.96 & \textbf{596} & 51MB \\
        \bottomrule
    \end{tabular}
    }
    \vspace{0.3cm}
    \caption{\small{\textbf{Quantitative comparisons on the Panoptic and Neural 3D Video datasets.} The best result is shown in bold, and the second-best is underlined. $^{\ast}$STG on the Neural 3D Video dataset is trained on 50-frame sequences and requires six models for evaluation.}}
    \label{tab:combined_results}
    \vspace{-0.4cm}
\end{table}

On the Panoptic dataset, we were unable to obtain reasonable results for the STG baseline~\cite{stg}, while 4D Gaussian~\cite{4dgs} produce very distorted images, as shown in Figure \ref{fig:Comparison}. This dataset demonstrates the strengths of explicit methods such as Dynamic 3DGS~\cite{luiten2023dynamic} and ours. While Dynamic 3DGS~\cite{luiten2023dynamic} effectively models complex motion and performs well on the Panoptic dataset, it is highly memory inefficient. In contrast, our model achieves similar performance with significantly lower storage. Figure~\ref{fig:Main2} shows a comparison with other methods.

\begin{table*}[t]
\centering
\small
\vspace{0.2cm}
\scalebox{0.9}{
\begin{tabular}{cccccccccccccc}\toprule
\multicolumn{3}{c}{Method \textbackslash Dataset} & \multicolumn{4}{c}{Basketball (Panoptic)}                 & \multicolumn{4}{c}{Cook Spinach (N3VD)}                   \\
\cmidrule(lr){1-3}\cmidrule(lr){4-7}\cmidrule(lr){8-11}
M       & Q       & I      & PSNR  & \#Gauss & Storage & FPS & PSNR & \#Gauss & Storage & FPS \\\midrule
\multicolumn{3}{c}{Dynamic 3DGS}                          & 28.2    & 349K  & 2161 MB   & 582 & 33.1 & 294K  & 3370 MB  & 472 \\
\checkmark          &             &            &  28.1  & 189K   & 1087 MB  & 750 & 32.9 & 59 K   & 674 MB  & 583 \\
\checkmark          & \checkmark         &           & 27.9  & 189K   & 299 MB   & 750 & 32.8 & 59 K   & 194 MB   & 583 \\
\checkmark          & \checkmark          & \checkmark & 27.9  & 189K   & 44 MB   & 750 & 32.7 & 59 K   & 53 MB   & 583 \\
\bottomrule
\end{tabular}
}
\vspace{0.3cm}
\caption{\small{\textbf{Ablation study on the proposed contributions.} `M', `Q' and `I' denote masking, quantization, interpolation, respectively. `\#Gauss' means the number of Gaussians.}}
\label{tab:ablation}
\vspace{-0.6cm}
\end{table*}

\vspace{-0.3cm}
\subsection{Ablation Studies}
\vspace{-0.05cm}

We conduct an ablation study to evaluate the effectiveness of each component in our method. Table~\ref{tab:ablation} presents results on two scenes from the Panoptic dataset~\cite{Joo_2019} and Neural 3D Video dataset~\cite{li2022neural3dvideosynthesis}, demonstrating how each step contributes to reducing storage. Our pruning strategy reduces the number of Gaussians by 2 and by 5 folds, followed by sensitivity-aware quantization, which compresses storage by 5 times, and keypoint interpolation, adding a further ~5 times reduction. Overall, TC3DGS achieves compression ratios of 49 and 64, respectively, with minimal impact on novel view synthesis, which is barely perceptible. Additional ablations are provided in the Supplementary.

\subsection{Mixed-Precision Quantization}
We ran experiments with sensitivity-aware quantization with different bit-ranges by varying $b_{\text{min}}$ and $b_{\text{max}}$ as well as using uniform bitwidth for all paramaters. We shown in the Table \ref{tab:quantization} that using adaptive bitwidth, the average bitwidth is lower than uniform quantization at 8 bits while image quality is similar. Whereas, compared to uniform quantization using lower bitwidth, our average bitwidth is slightly higher while image quality is improved considerably.

\begin{table}[h]
\vspace{-0.08cm}
\caption{\textbf{Comparison of Sensitivity-Aware and Uniform Bit Quantization.} Our sensitivity-aware [4,8]-bit quantization achieves PSNR comparable to 8-bit precision while providing compression similar to uniform 5-bit quantization.}
\label{tab:quantization}
\vspace{0.4cm}
\centering
\scalebox{0.89}{
\begin{tabular}{c|cccc}
\toprule
\textbf{Bit-precision} & \textit{PSNR} & \textit{SSIM} & \textit{LPIPS} & \textit{Compression}\\
\midrule
\midrule
\textit{uniform 4 bit} & 25.1 & 0.81 & 0.32 & 8x \\
\textit{uniform 5 bit}  & 25.8 & 0.84 & 0.28 & 6.4x \\
\textit{uniform 6 bit}  & 26.1 & 0.86 & 0.26 & 5.3x \\
\textit{uniform 8 bit}   & 28.1 & 0.90 & 0.20 & 4x \\
\textit{Ours [5,8]} & 28.0 & 0.90 & 0.20 &  5.6x\\
\textit{Ours [4,8]} & 27.9 & 0.90 & 0.20 & 6.3x \\
\textit{Ours [3,8]} & 26.2 & 0.84 & 0.28 & 7.3x \\
\bottomrule
\end{tabular}
}

\end{table}

\vspace{-0.4cm}
\section{Conclusion}
\vspace{-0.18cm}
We have introduced Temporal Compressed 3D Gaussian Splatting (TC3DGS), a novel framework designed to achieve memory-efficient and high-speed reconstruction of dynamic scenes. Our approach achieves up to 67x compression and up to three times faster rendering while maintaining high fidelity in complex motion scenarios. Through selective temporal pruning and gradient-based quantization, TC3DGS significantly reduces memory usage with minimal impact on visual quality. While our method is effective, limitations remain in aggressively compressing position data and handling new objects that enter the scene mid-sequence. Future work will focus on bridging the gap between spatio-temporal methods and storage-efficient dynamic scene reconstruction, improving adaptability and further reducing storage requirements for complex, evolving environments.

\bibliography{egbib}
\newpage
\setcounter{page}{1}
\maketitle

\begin{center}
\vspace{-0.5cm}
    Supplementary
\end{center}


\section{Implementation Details}
For our experiments, we closely follow the hyperparameters outlined in Dynamic 3DGS \cite{luiten2023dynamic}. Specifically, for the compression strategy, we set both the mask weight parameter $\lambda_{\text{mask}}$ and the masking consistency loss parameter $\lambda_{\text{mask-cons}}$ to $0.01$ in our primary experiments. In terms of quantization, we use Gaussian parameters with bin sizes of $b_{\text{min}} = 4$ and $b_{\text{max}} = 8$, while positional data is quantized to 16-bit precision to ensure spatial accuracy. For the learnable quantization step size, we use a $0.02$ learning rate.  Quantization is applied after 6000 iterations in the first scene for all experiments. For keypoint interpolation, we set the tolerance to $\tau = 1 \times 10^{-5}$ for the Panoptic dataset and $\tau = 1 \times 10^{-7}$ otherwise. Moreover, we restrict the maximum number of keypoints to $\text{max}_{\text{kp}} = 30$ for the Panoptic dataset and $\text{max}_{\text{kp}} = 60$ otherwise, which provides an effective balance between compression and temporal trajectory accuracy. Additional ablation studies with varying values of hyperparameters are provided in the supplementary.

\section{Additional Ablation Experiments}
\label{sec:additional_datasets}
\begin{figure}[h]
    \centering
    \begin{tabular}{c@{\hspace{-0.5em}}c@{\hspace{-0.5em}}c@{\hspace{-0.5em}}c}
 
        \vspace{-0.5em}
        \includegraphics[width=0.25\textwidth]{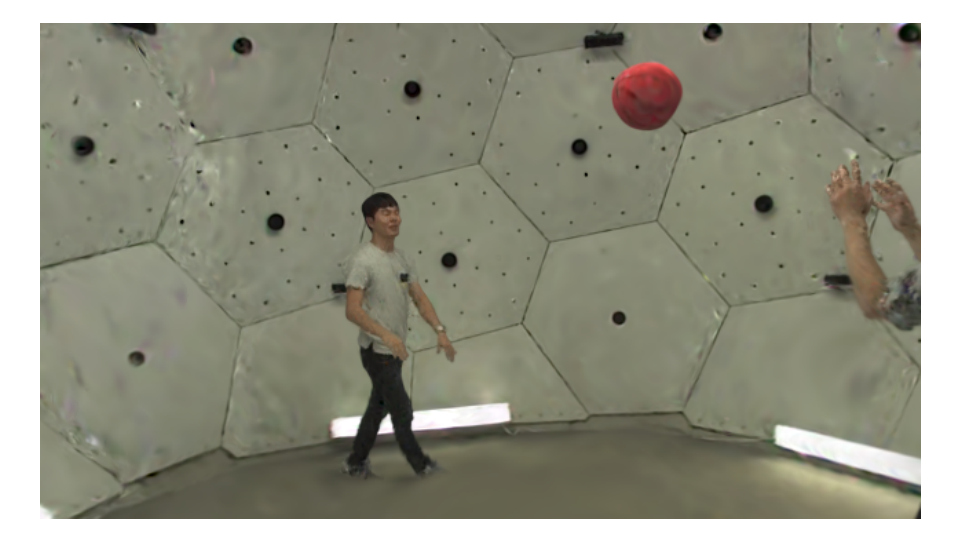} & 
        \includegraphics[width=0.25\textwidth]{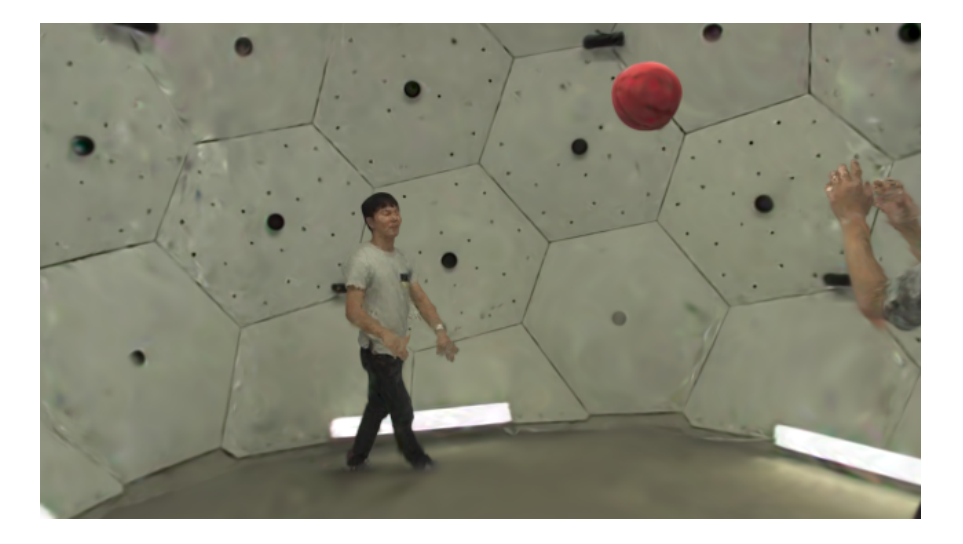} & 
        \includegraphics[width=0.25\textwidth]{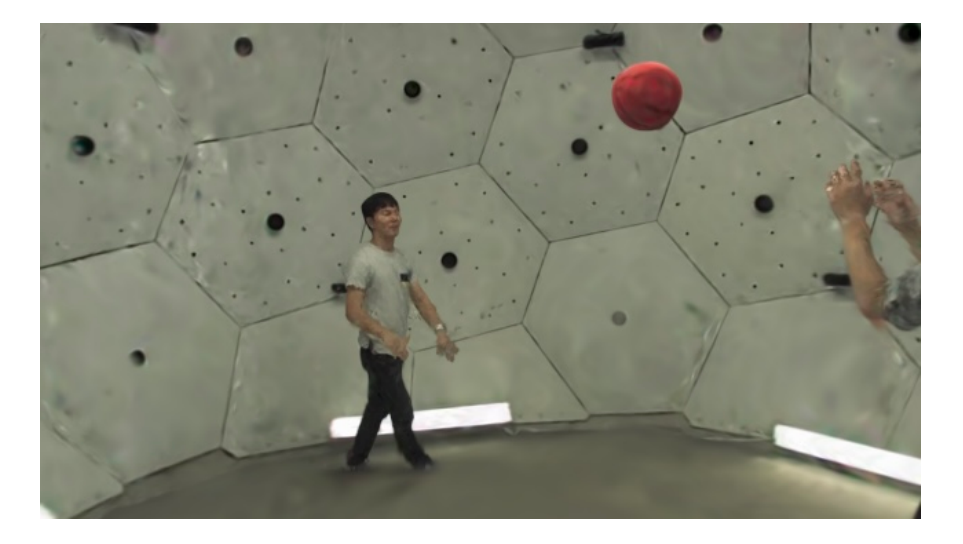} & 
        \includegraphics[width=0.25\textwidth]{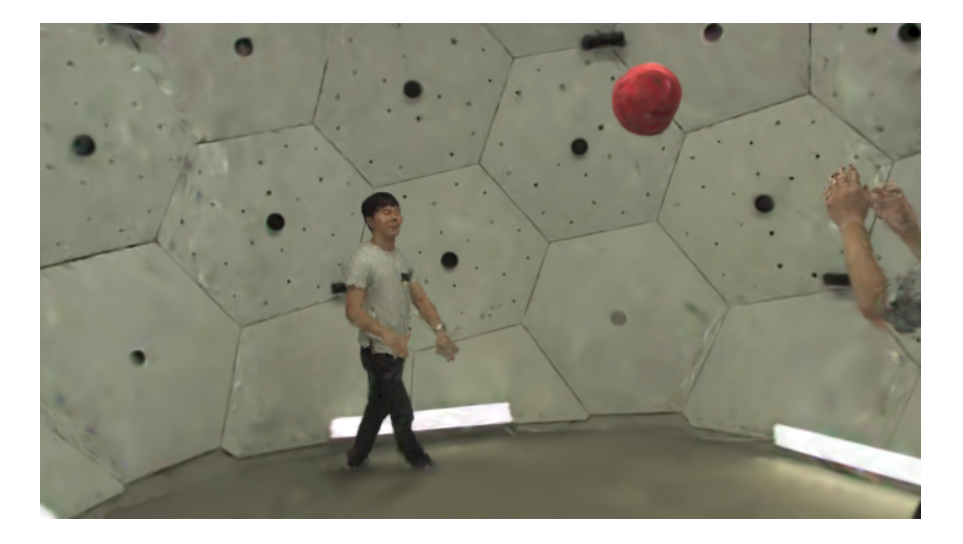} \\ 

        \vspace{-0.5em}
        \includegraphics[width=0.25\textwidth]{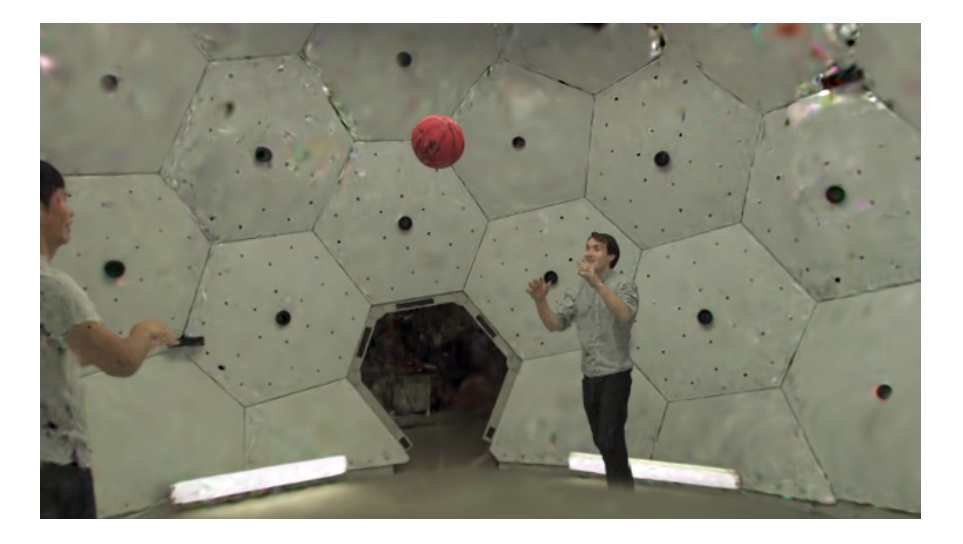} & 
        \includegraphics[width=0.25\textwidth]{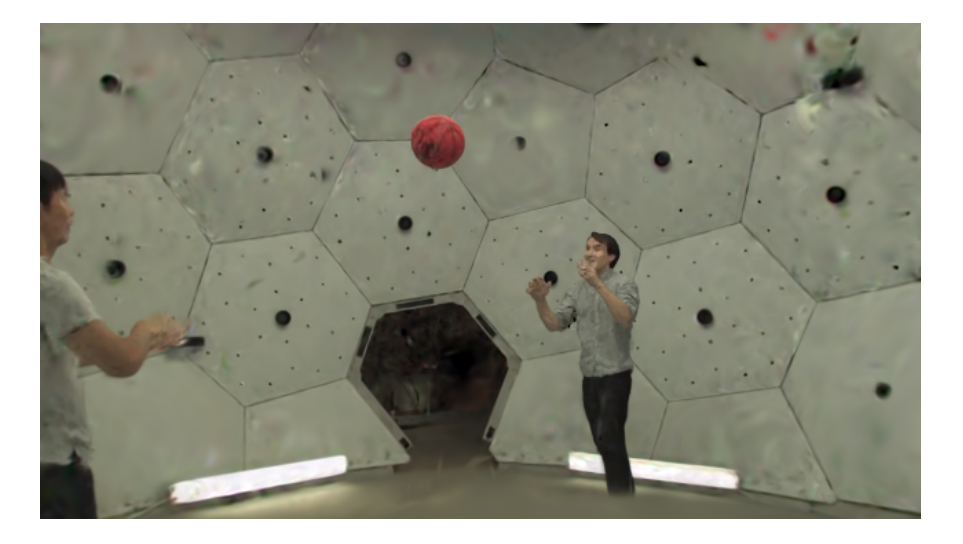} & 
        \includegraphics[width=0.25\textwidth]{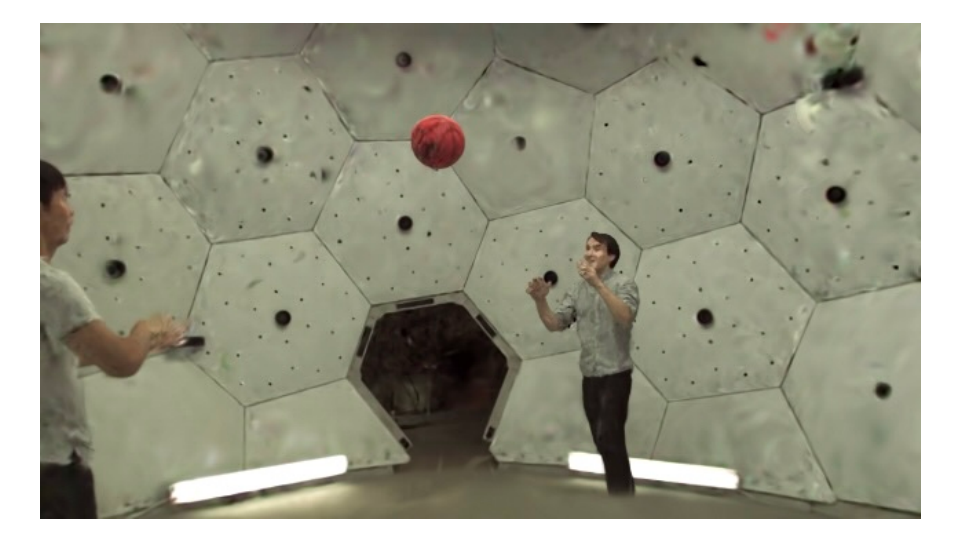} & 
        \includegraphics[width=0.25\textwidth]{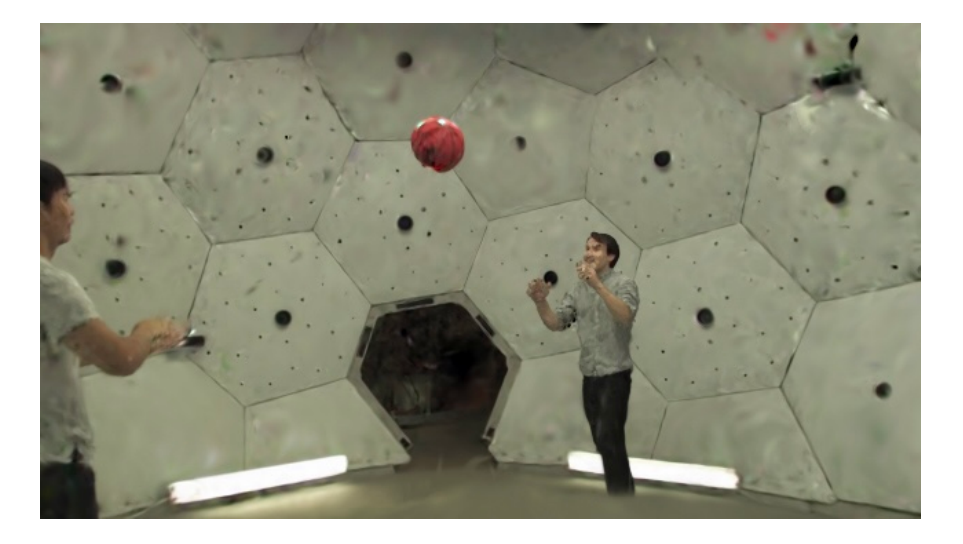} \\ 

        \vspace{-0.2em} 
        \includegraphics[width=0.25\textwidth]{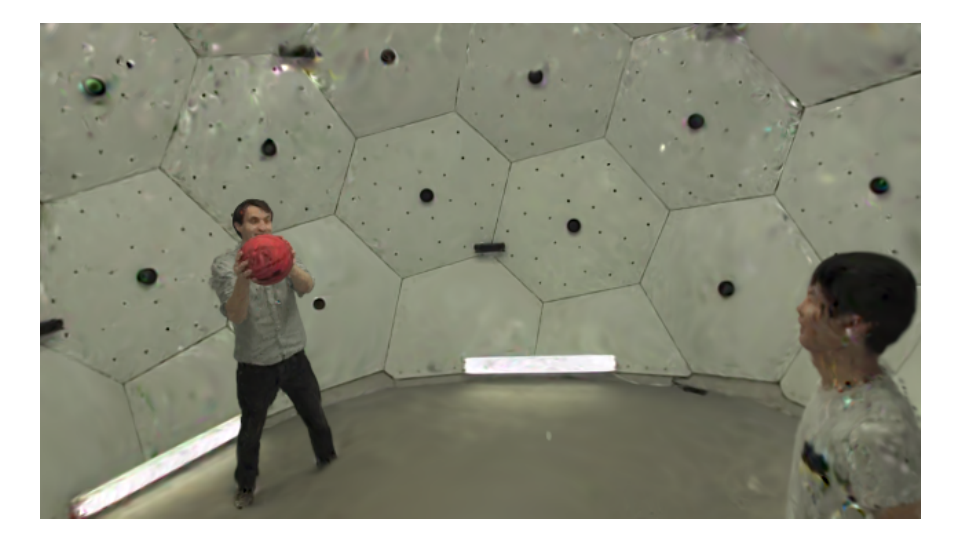} & 
        \includegraphics[width=0.25\textwidth]{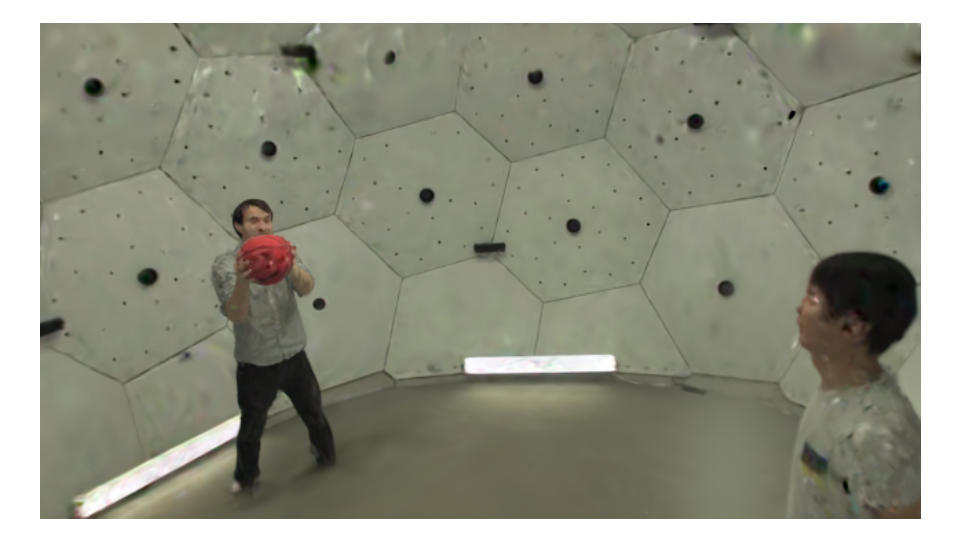} & 
        \includegraphics[width=0.25\textwidth]{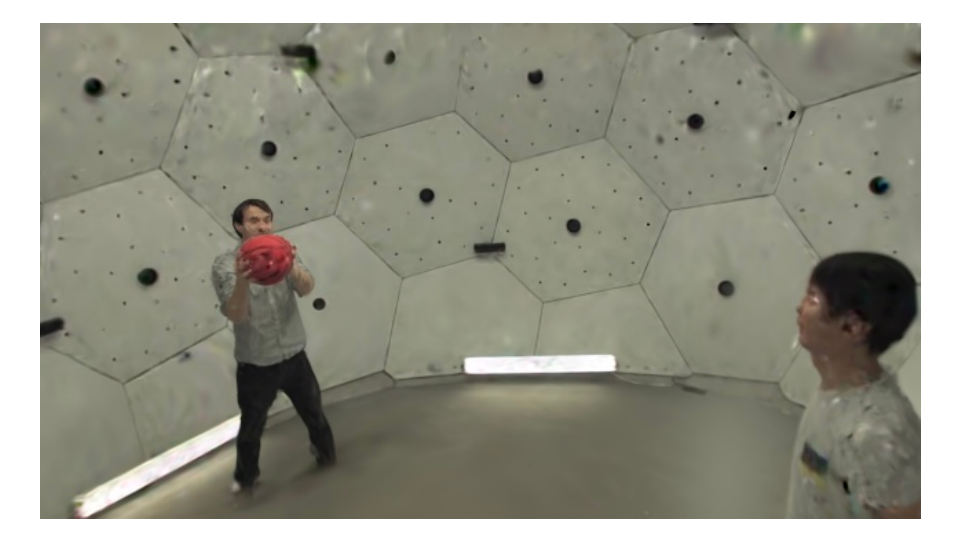} & 
        \includegraphics[width=0.25\textwidth]{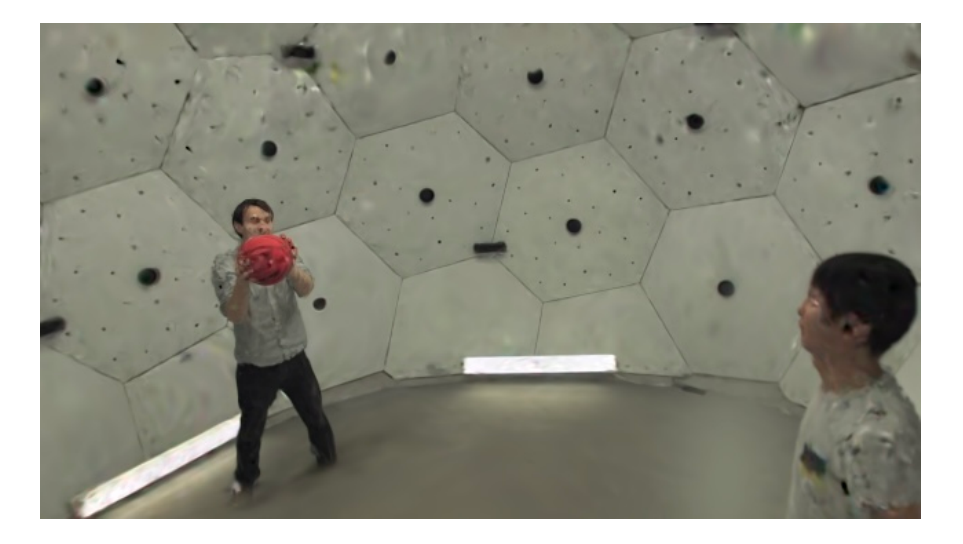} \\ 

        \small{Dynamic 3DGS~\cite{luiten2023dynamic}} & \small{Pruned} & \small{Quantized + Pruned} & \small{Quantized + Pruned + KPI} \\
    \end{tabular}
    \vspace{0.3cm}
    \caption{\textbf{Ablation Study on Panoptic Sports Dataset} }
    \label{fig:Ablation}
\end{figure}

We conducted several experiments to better understand hyperparameters of our method as shown in Figure~\ref{fig:Ablation}.

\subsection{Masking and Mask Consistency}

We vary the weight of the mask loss $\lambda_{\text{mask}}$ and the masking consistency loss $\lambda_{\text{mask-cons}}$ on the first 50 frames of the \textit{Basketball} scene from the Panoptic Sports dataset. The results, shown in Table \ref{tab:mask_ablation_results}, indicate that increasing the weight of these parameters leads to greater pruning of Gaussians but results in reduced image quality. We found $\lambda_{\text{mask}} = \lambda_{\text{mask-cons}} = 0.01$ to be an effective balance between maintaining image quality and minimizing model size.

\begin{table}
\vspace{-0.3cm}
\begin{subtable}[t]{0.4\linewidth} 
\centering
\caption{SSIM $\uparrow$}
\label{tab:mask_ablation_ssim}
\begin{tabular}{c|cccc}
\toprule
& \multicolumn{4}{c}{\textbf{Mask Consistency Loss}} \\
\midrule
\textbf{Loss} & \textit{0} & \textit{0.01} & \textit{0.05} & \textit{0.1} \\
\midrule
\textit{0.005} & 0.909 & 0.911 & 0.911 & 0.907 \\
\textit{0.01}  & 0.913 & \textbf{0.914} & 0.911 & 0.909 \\
\textit{0.05}  & 0.904 & 0.901 & 0.899 & 0.897 \\
\textit{0.1}   & 0.894 & 0.896 & 0.895 & 0.897 \\
\bottomrule
\end{tabular}
\end{subtable}
\hspace{1.8cm}
\begin{subtable}[t]{0.4\textwidth}
\vspace{0.02cm}
\centering
\caption{PSNR $\uparrow$}
\label{tab:mask_ablation_psnr}
\begin{tabular}{c|cccc}
\toprule
& \multicolumn{4}{c}{\textbf{Mask Consistency Loss}} \\
\midrule
\textbf{Loss} & \textit{0} & \textit{0.01} & \textit{0.05} & \textit{0.1} \\
\midrule
\textit{0.005} & 28.0 & 27.8 & 28.2 & 26.8 \\
\textit{0.01}  & 28.1 & \textbf{28.3} & 26.8 & 26.9 \\
\textit{0.05}  & 27.2 & 26.6 & 26.0 & 26.0 \\
\textit{0.1}   & 27.1 & 27.4 & 26.6 & 26.7 \\
\bottomrule
\end{tabular}
\end{subtable}

\vspace{0.5em} 

\begin{subtable}[t]{0.4\textwidth}
\centering
\caption{LPIPS $\downarrow$}
\label{tab:mask_ablation_lpips}
\begin{tabular}{c|cccc}
\toprule
& \multicolumn{4}{c}{\textbf{Mask Consistency Loss}} \\
\midrule
\textbf{Loss} & \textit{0} & \textit{0.01} & \textit{0.05} & \textit{0.1} \\
\midrule
\textit{0.005} & 0.180 & 0.177 & 0.178 & 0.182 \\
\textit{0.01}  & \textbf{0.174} & 0.176 & 0.175 & 0.183 \\
\textit{0.05}  & 0.205 & 0.209 & 0.214 & 0.217 \\
\textit{0.1}   & 0.235 & 0.229 & 0.234 & 0.234 \\
\bottomrule
\end{tabular}
\end{subtable}
\hspace{1.8cm}
\begin{subtable}[t]{0.4\textwidth}
\vspace{0.02cm}
\centering
\caption{Number of Gaussians $\downarrow$}
\label{tab:mask_ablation_gaussians}
\begin{tabular}{c|rrrr}
\toprule
& \multicolumn{4}{c}{\textbf{Mask Consistency Loss}} \\
\midrule
\textbf{Loss} & \textit{0} & \textit{0.01} & \textit{0.05} & \textit{0.1} \\
\midrule
\textit{0.005} & 208k & 156k & 143k & 141k \\
\textit{0.01}  & 154k & 113k & 98k  & 94k \\
\textit{0.05}  & 50k  & 47k  & 30k  & 26k \\
\textit{0.1}   & 29k  & 31k  & 19k  & \textbf{16k} \\
\bottomrule
\end{tabular}
\end{subtable}
\vspace{0.35cm}
\caption{{\textbf{Impact of Mask Loss and Mask Consistency Loss.} Increasing the weight of these masks leads to more aggressive pruning of Gaussians, but at the cost of rendering quality. Setting $\lambda_{mask} = 0.01$ and $\lambda_{mask-cons} = 0.01$ provides a good trade-off between quality and storage efficiency.}}
\label{tab:mask_ablation_results}
\end{table}

\begin{figure}
    \vspace{-1cm}
    \centering
    \vspace{0.2cm}
    \begin{minipage}{0.46\linewidth}
        \centering
        \includegraphics[width=0.9\linewidth]{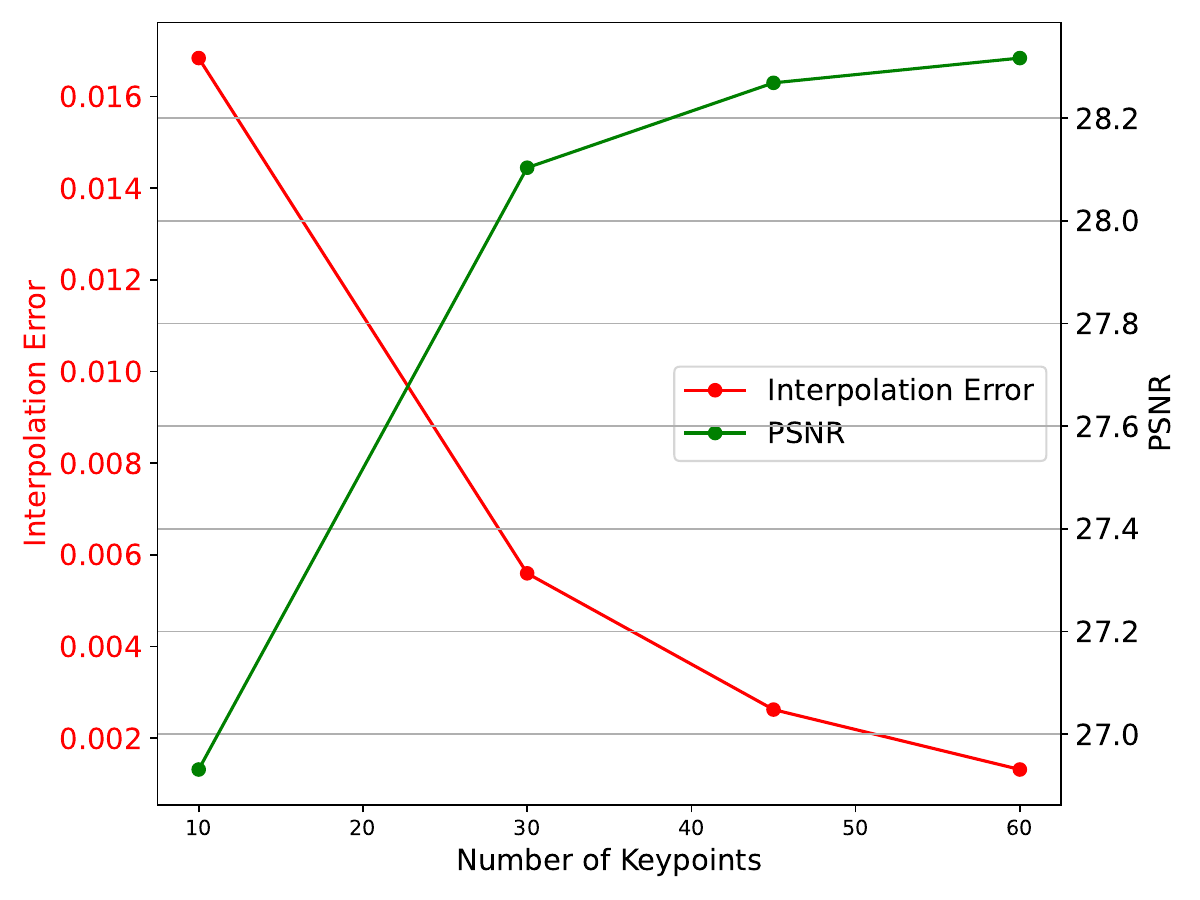}
        \subcaption{Interpolation Error and PSNR vs Maximum Keypoints}
    \end{minipage}
    \hfill
    \begin{minipage}{0.46\linewidth}
        \centering
        \vspace{0.45cm}\includegraphics[width=0.9\linewidth]{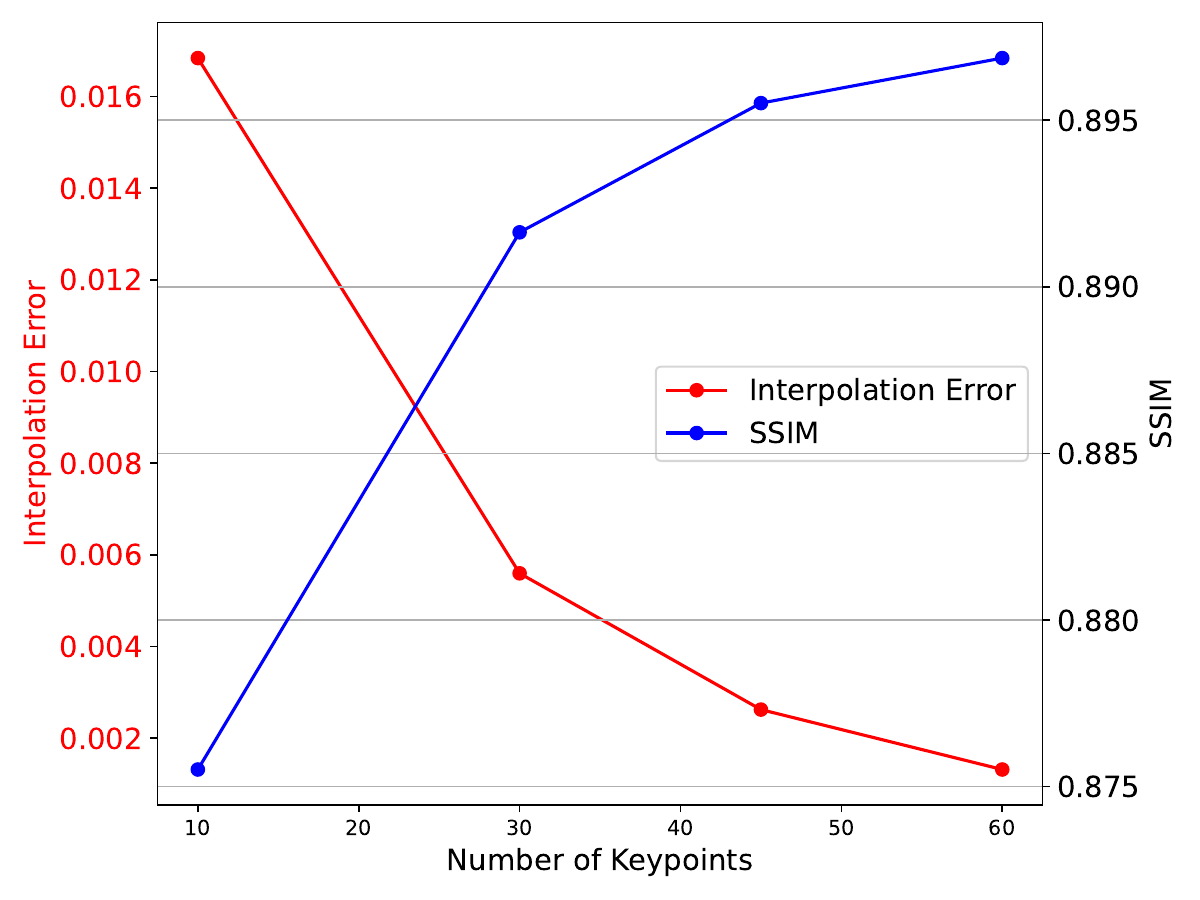}
        \subcaption{Interpolation Error and SSIM vs Maximum Keypoints\\}
    \end{minipage}\\
    \vspace{0.3cm}
    \caption{{\textbf{Impact of increasing Maximum Keypoints} The figure illustrates the trend of interpolation error and PSNR/SSIM as we increase $max_{kp}$.  It highlights that the rate of change in interpolation error and PSNR/SSIM diminishes as we increase $max_{kp}$}}
    \label{fig:keypoint_error_graphs}
\end{figure}

\begin{figure}

    \centering
    \begin{minipage}{0.48\linewidth}
        \centering
        \includegraphics[width=\linewidth]{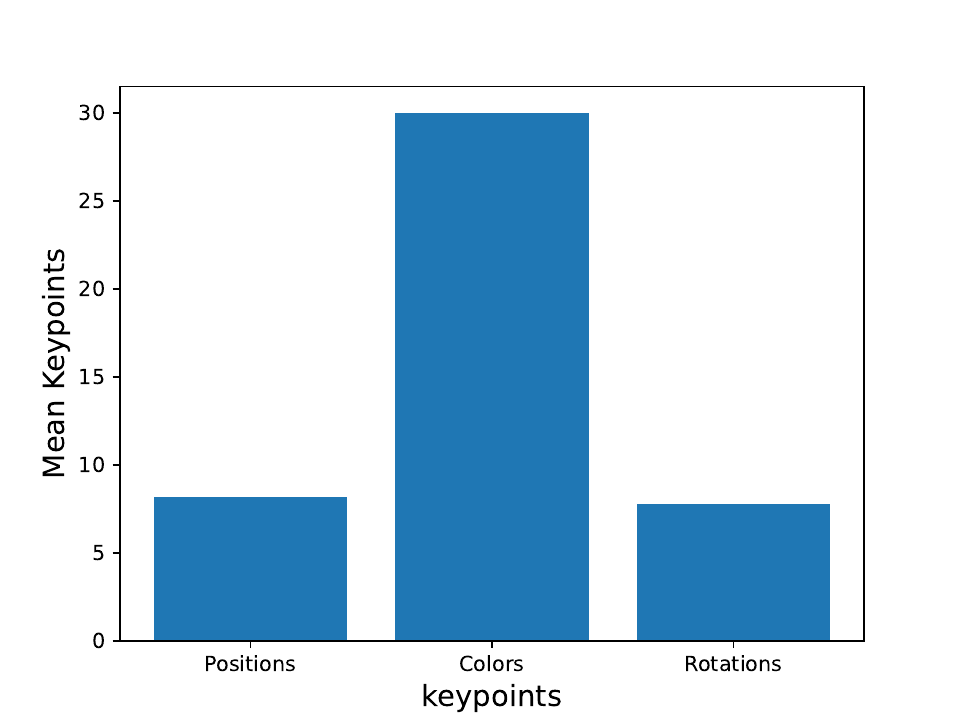}
        \subcaption{Mean number of Keypoints}
    \end{minipage}
    \hfill
    \begin{minipage}{0.48\linewidth}
        \centering
        \includegraphics[width=\linewidth]{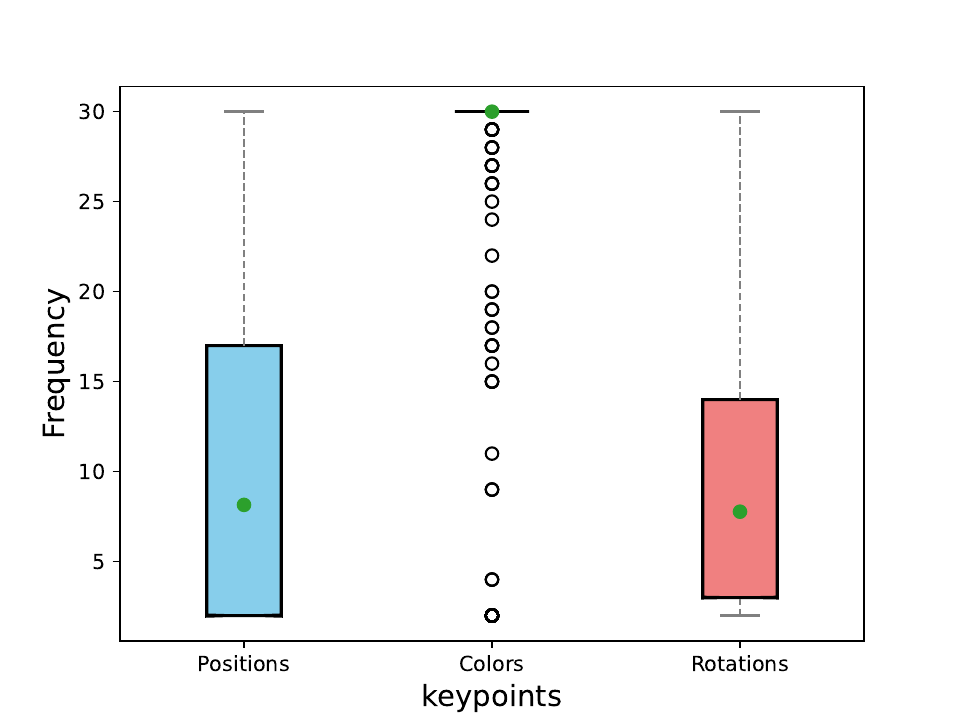}
        \subcaption{Distribution of Keypoints\\}
    \end{minipage}
    \vspace{0.3cm}
    \caption{\textbf{Distribution of Keypoints Across Different Parameters.} The figure illustrates the mean number of keypoints for Gaussian parameters. It highlights that only color parameters reach $max_{kp}$ saturation, while positions and rotations can be effectively estimated with fewer keypoints.}
    \label{fig:keypoint_graphs}
\end{figure}

\subsection{Keypoint Interpolation}
We experimented with varying the hyperparameters for keypoint interpolation, specifically the maximum number of keypoints, $max_{kp}$. As shown in Figure \ref{fig:keypoint_error_graphs}, increasing $max_{kp}$ reduces compression error and improves image quality. Conversely, increasing $\tau$ allows for greater error tolerance, leading to a decrease in image quality.

The experiments summarized in Figure \ref{fig:keypoint_error_graphs} were conducted on 150 frames of the \textit{Basketball} scene with masking parameters set to $\lambda_{\text{mask}} = \lambda_{\text{mask-cons}} = 0.01$ and fixed $tau = 1 \times e^{-6}$. We observed that overly relaxing $\tau$ leads to underfitted keypoints, resulting in compromised rendering quality. On the other hand, increasing $max_{kp}$ beyond a certain threshold results in diminishing returns, where interpolation error continues to decrease, but the improvement in image quality becomes marginal.

Additionally, Figure \ref{fig:keypoint_graphs} illustrates that only the color parameters saturate the maximum keypoints, whereas positions and rotations can be adequately approximated with fewer keypoints. 

We selected the hyperparameters to strike a balance between rendering quality and storage efficiency. For the Panoptic Sports dataset, we used $max_{kp} = 30$ and $\tau = 1 \times 10^{-5}$. For the Neural 3D Video dataset, which consists of 300-frame sequences, we increased $max_{kp}$ to 60 to accommodate the longer sequence length. In the case of the Technicolor dataset, despite its long sequences, we matched the experimental setup by using 50 timesteps per scene and reduced $max_{kp}$ to 15.

\section{Detailed Results for Panoptic and Neural 3D Video Dataset}
\label{sec:additional_datasets}
We present detailed results for the two datasets discussed in Section 5.1 of our paper, summarized in Table \ref{tab:panoptic_perscene} and \ref{tab:n3d_perscene}. These results include various metrics evaluated for each scene in both datasets, offering a granular view of performance. Additionally, we provide supplementary visualizations for the Neural 3D Video dataset, which further demonstrate the effectiveness of our method in capturing dynamic scenes with high fidelity. Finally, we present an ablation study over the main elements of our method in \Cref{fig:Ablation}.
\begin{table}[h!]
\caption{
\textbf{Per-scene quantitative comparisons on the Panoptic Sports Dataset~\cite{Joo_2019}. }
}
\label{tab:panoptic_perscene}
\vspace{0.3cm}
\centering
\scalebox{0.9}{
\begin{tabular}{l|l|llllll}
\toprule
Method & Avg. & \textit{Basket.} & \textit{Juggle} & \textit{Boxes} & \textit{Softball} & \textit{Tennis} & \textit{Football}  \\
\midrule
\midrule
\multicolumn{2}{l}{\textbf{PSNR$\uparrow$}} \\
\midrule
4DGaussians~\cite{4dgs} & $27.2$ & $26.68$ & $27.47$ & $27.02$ & $27.44$ & $27.31$ & $27.39$ \\
STG~\cite{stg} & $20.45$ & $21.60$ & $19.93$ & $20.65$ & $19.44$ & $20.82$ & $20.23$ \\
D. 3DGS~\cite{luiten2023dynamic} & $\textbf{28.70}$ & $\textbf{28.22}$ & $\textbf{29.48}$ & $\textbf{29.46}$ & $\textbf{28.43}$ & $\textbf{28.11}$ & $28.49$ \\
HiCom~\cite{hicom2024} & $26.59$ & $26.35$ & $27.15$ & $25.09$ & $28.38$ & $25.91$ & $\textbf{28.62}$ \\
\midrule
Ours & $27.81$ & $28.02$ & $28.65$ &$28.28$& $27.96$ & $25.97$ & $28.00$ \\
\midrule
\midrule
\multicolumn{2}{l}{\textbf{SSIM$\uparrow$}} \\
\midrule
4DGaussians~\cite{4dgs} & $\textbf{0.91}$ &$0.90$ & $\textbf{0.92}$ & $\textbf{0.91}$ & $\textbf{0.92}$ & $\textbf{0.92}$ & $\textbf{0.92}$ \\
STG~\cite{stg} & $0.79$ & $0.78$ & $0.78$ & $0.79$ & $0.79$ & $0.78$ & $0.78$ \\
D. 3DGS~\cite{luiten2023dynamic} & $\textbf{0.91}$ & $\textbf{0.91}$ & $\textbf{0.92}$ & $\textbf{0.91}$ & $0.91$ & $0.91$ & $0.91$ \\
HiCoM~\cite{hicom2024} & $0.87$ & $0.82$ & $0.88$ & $0.86$ & $0.90$ & $0.86$ & $0.91$ \\
\midrule
Ours & $0.89$ & $0.89$ & $0.90$ &$0.89$& $0.89$ & $0.89$ & $0.89$ \\
\midrule
\midrule
\multicolumn{2}{l}{\textbf{SIZE(MB)$\downarrow$}} \\
\midrule
4DGaussians~\cite{4dgs} & $63$ & $66$ & $59$ & $63$ & $57$ & $72$ & $65$ \\
STG~\cite{stg} & $\textbf{19}$ & $\textbf{19}$ & $\textbf{19}$ & $\textbf{20}$ & $\textbf{19}$ & $\textbf{22}$ & $\textbf{19}$ \\
D. 3DGS~\cite{luiten2023dynamic} & $2008$ & $2161$ & $1935$ & $2021$ & $2021$ & $1915$ & $2000$ \\
HiCoM~\cite{hicom2024} & $71$ & $71$ & $71$ & $76$ & $70$ & $69$ & $70$ \\
\midrule
Ours& $49$ & $46$ & $32$ & $44$ & $41$ & $48$ & $34$ \\
\bottomrule
\end{tabular}
}
\end{table}
\begin{table}
\caption{
\textbf{Per-scene quantitative comparisons on the Neural 3D Video Dataset~\cite{li2022neural3dvideosynthesis}.}
Some methods only report part of the scenes. 
$^1$ only includes the \textit{Flame Salmon} scene.
$^2$ excludes the \textit{Coffee Martini} scene.
``-" denotes results that are unavailable in prior work. $^{\ast}$STG on the Neural 3D Video dataset is trained on 50-frame sequences and requires six models for evaluation.
}
\vspace{0.3cm}
\label{tab:n3d_perscene}
\centering
\resizebox{1\linewidth}{!}{%
\begin{tabular}{l|l|llllll}
\toprule
Method & Avg. & \textit{Coffee Martini} & \textit{Cook Spinach} & \textit{Cut Roasted Beef} & \textit{Flame Salmon} & \textit{Flame Steak} & \textit{Sear Steak} \\
\midrule
\midrule
\multicolumn{2}{l}{\textbf{PSNR$\uparrow$}} \\
\midrule
Neural Volumes~\cite{Lombardi2019} $^1$ & $22.80$ & - & - & - & $22.80$ & - & - \\ 
LLFF~\cite{mildenhall2019llff} $^1$ & $23.24$ & - & - & - & $23.24$ & - & - \\ 
DyNeRF~\cite{li2022neural3dvideosynthesis} $^1$ & $29.58$ & - & - & - & $29.58$ & - & - \\ 
HexPlane~\cite{Cao2023HexPlane} $^2$ & $31.71$ & - & $32.04$ & $32.55$ & $29.47$ & $32.08$ & $32.39$ \\
NeRFPlayer~\cite{song2023nerfplayer} & $30.69$ & $\textbf{31.53}$ & $30.56$ & $29.35$ & $31.65$ & $31.93$ & $29.13$ \\
HyperReel~\cite{attal2023hyperreel} & $31.10$ & $28.37$ & $32.30$ & $32.92$ & $28.26$ & $32.20$ & $32.57$ \\
K-Planes~\cite{kplanes_2023} & $31.63$ & $29.99$ & $32.60$ & $31.82$ & $30.44$ & $32.38$ & $32.52$ \\
MixVoxels-L~\cite{Wang2023ICCV} & $31.34$ & $29.63$ & $32.25$ & $32.40$ & $29.81$ & $31.83$ & $32.10$ \\
MixVoxels-X~\cite{Wang2023ICCV} & $31.73$ & $30.39$ & $32.31$ & $32.63$ & $\textbf{30.60}$ & $32.10$ & $32.33$ \\
4D Gaussians~\cite{4dgs} & $30.67$ & $27.34$ & $32.46$ & $32.90$ & $29.20$ & $32.51$ & $32.49$ \\
Dynamic 3DGS~\cite{luiten2023dynamic} & $30.97$ & $27.32$ & $32.97$ & $31.75$ & $27.26$ & $33.24$ & $33.68$ \\
STG~\cite{stg} & $\textbf{32.05}$ & $28.61$ & $\textbf{33.18}$ &$33.52$& $29.48$ & $\textbf{33.64}$ & $\textbf{33.89}$ \\
\midrule
Ours  & $30.96$ & $27.42$ & $32.72$ & $\textbf{33.61}$ & $27.12$ & $32.48$ & $33.04$ \\
\midrule
\midrule

\multicolumn{2}{l}{\textbf{LPIPS$\downarrow$}} \\
\midrule
Neural Volumes~\cite{Lombardi2019} $^1$ & $0.295$ & - & - & - & $0.295$ & - & - \\ 
LLFF~\cite{mildenhall2019llff} $^1$ & $0.235$ & - & - & - & $0.235$ & - & - \\ 
DyNeRF~\cite{li2022neural3dvideosynthesis} $^1$ & $0.083$ & - & - & - & $0.083$ & - & - \\ 
HexPlane~\cite{Cao2023HexPlane} $^2$ & $0.075$ & - & $0.082$ & $0.080$ & $0.078$ & $0.066$ & $0.070$ \\
NeRFPlayer~\cite{song2023nerfplayer} & $0.111$ & $0.085$ & $0.113$ & $0.144$ & $0.098$ & $0.088$ & $0.138$ \\
HyperReel~\cite{attal2023hyperreel} & $0.096$ & $0.127$ & $0.089$ & $0.084$ & $0.136$ & $0.078$ & $0.077$ \\
MixVoxels-L~\cite{Wang2023ICCV} & $0.096$ & $0.106$ & $0.099$ & $0.088$ & $0.116$ & $0.088$ & $0.080$ \\
MixVoxels-X~\cite{Wang2023ICCV} & $0.064$ & $0.081$ & $0.062$ & $0.057$ & $0.078$ & $0.051$ & $0.053$ \\
Dynamic 3DGS~\cite{luiten2023dynamic} & $0.082$ & $0.122$ & $0.077$ & $0.087$ & $0.09$ & $0.072$ & $0.068$ \\
$^{\ast}$STG~\cite{stg}& $\textbf{0.044}$ & $\textbf{0.069}$ & $\textbf{0.037}$ & $\textbf{0.036}$ & $\textbf{0.063}$ & $\textbf{0.029}$ & $\textbf{0.030}$ \\
\midrule    
Ours & $0.089$ & $0.129$ & $0.081$ & $0.088$ & $0.094$ & $0.075$ & $0.071$  \\
\bottomrule
\end{tabular}
}
\end{table}

\section{Technicolor Dataset~\cite{technicolor}}
\begin{figure*}[t]

    \centering
    \begin{minipage}{0.49\linewidth}
        \centering
        \includegraphics[width=\linewidth]{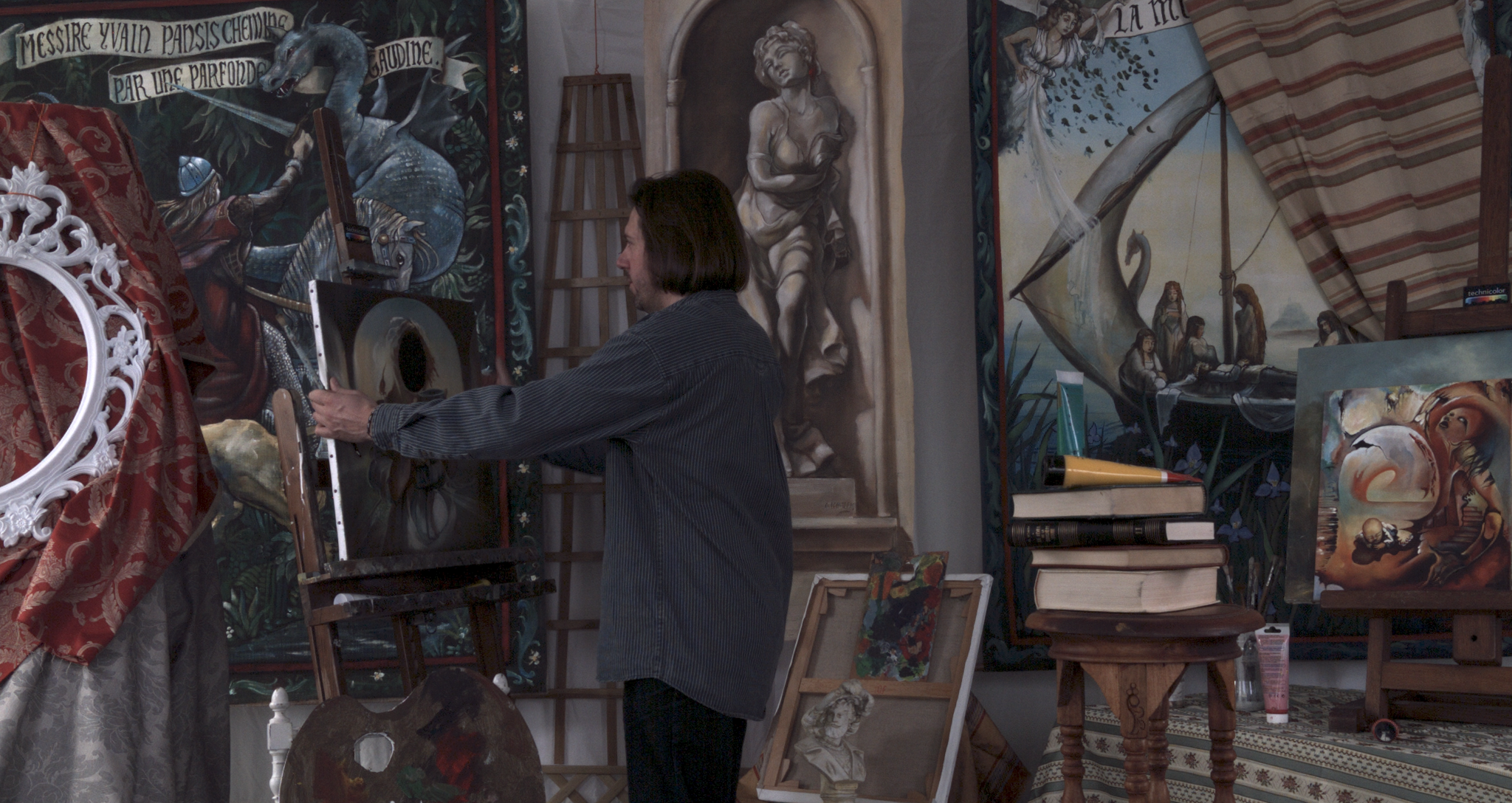}
        \includegraphics[width=\linewidth]{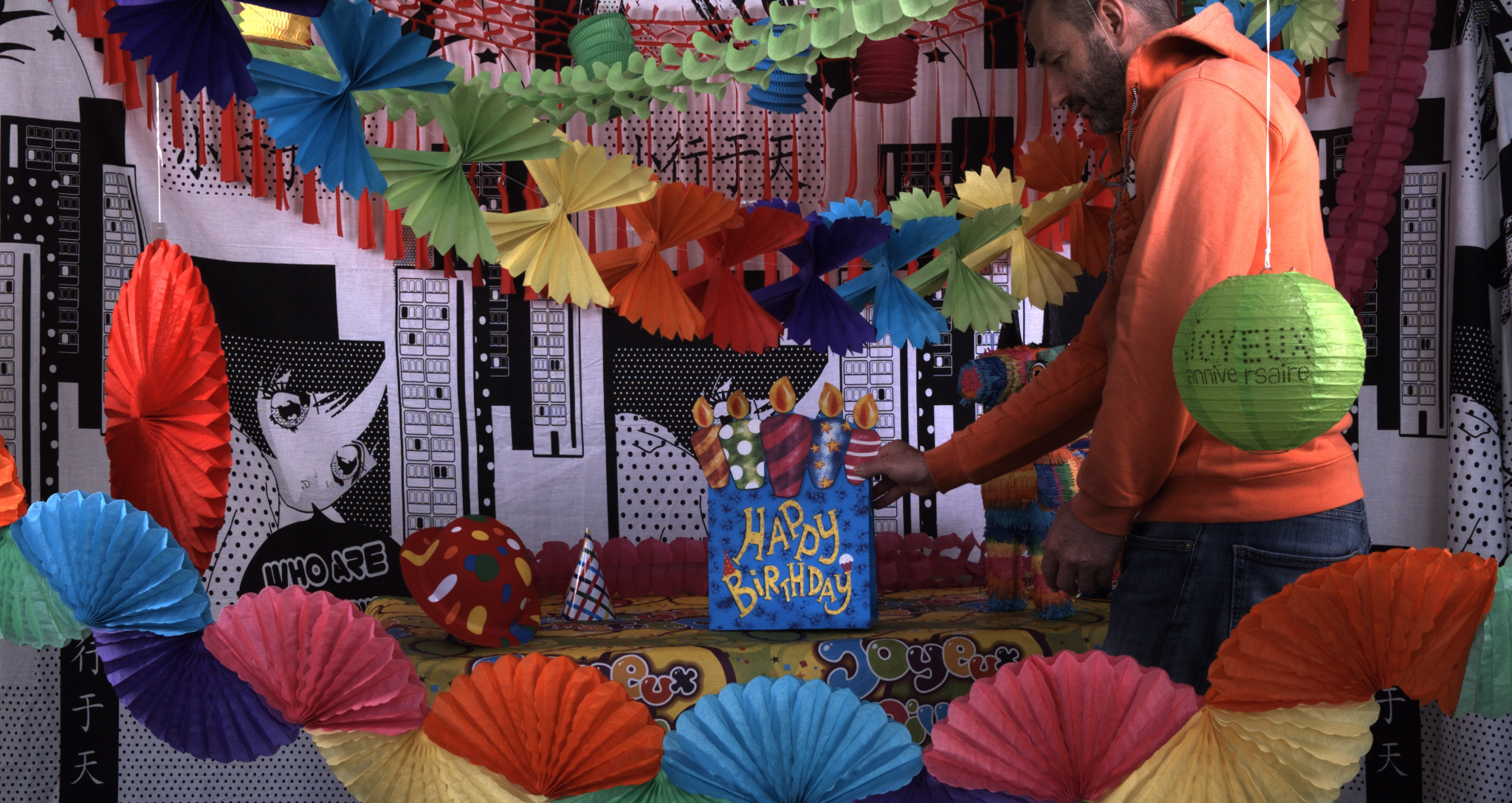}
        \includegraphics[width=\linewidth]{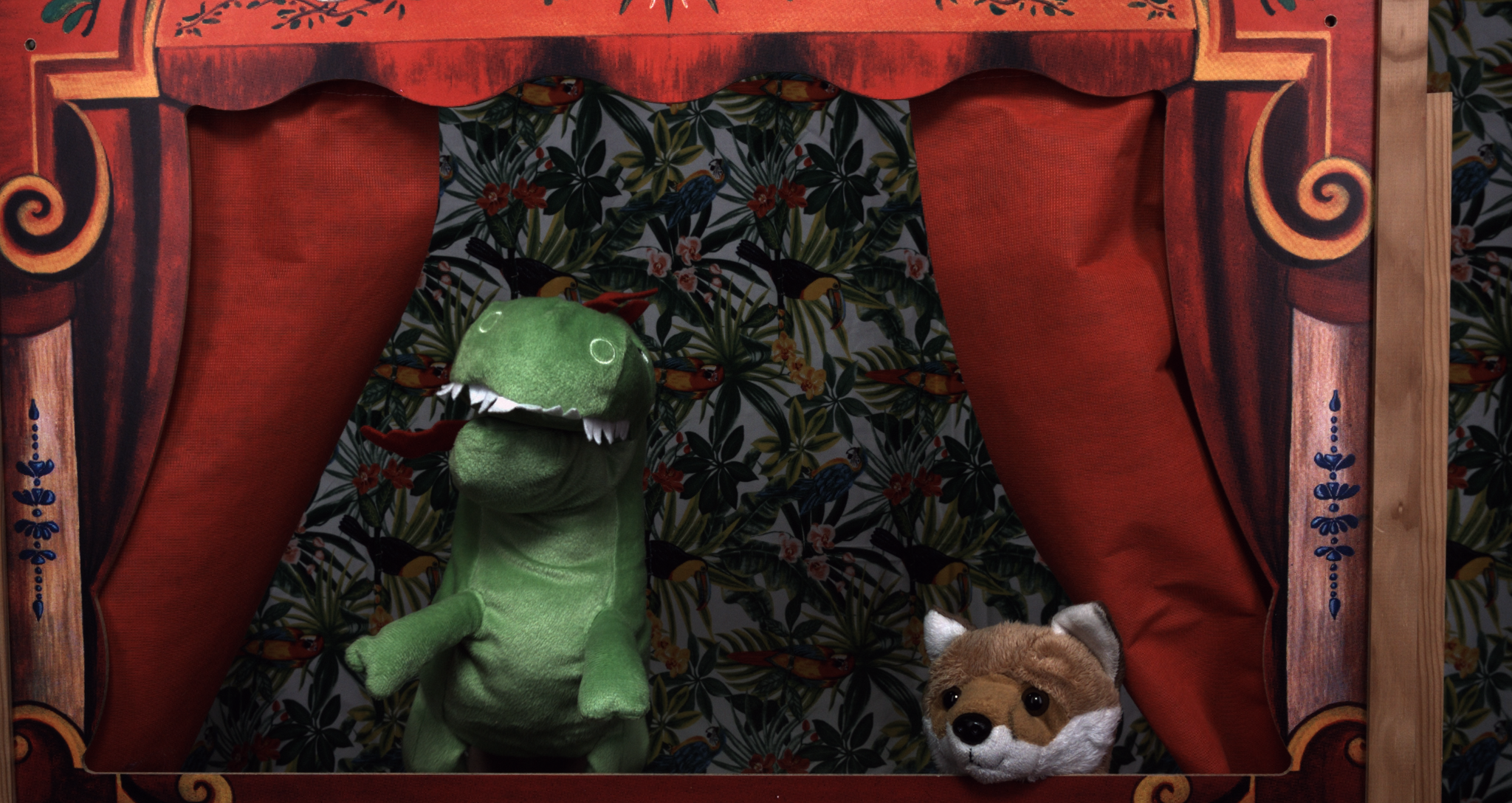}
        \includegraphics[width=\linewidth]{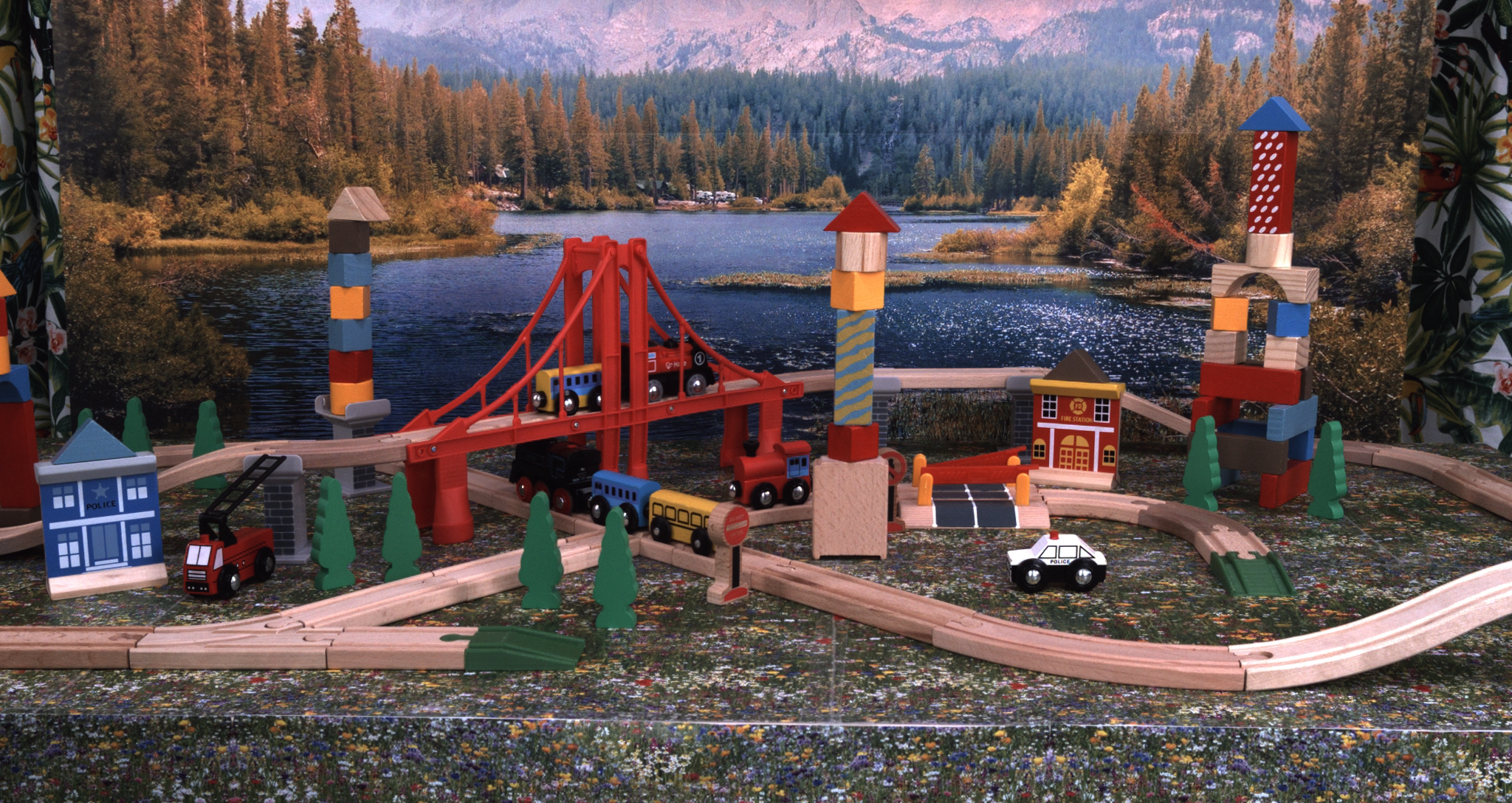}

        \subcaption{Ground-Truth}
    \end{minipage}
    \hfill
    \begin{minipage}{0.49\linewidth}
        \centering
        \includegraphics[width=\linewidth]{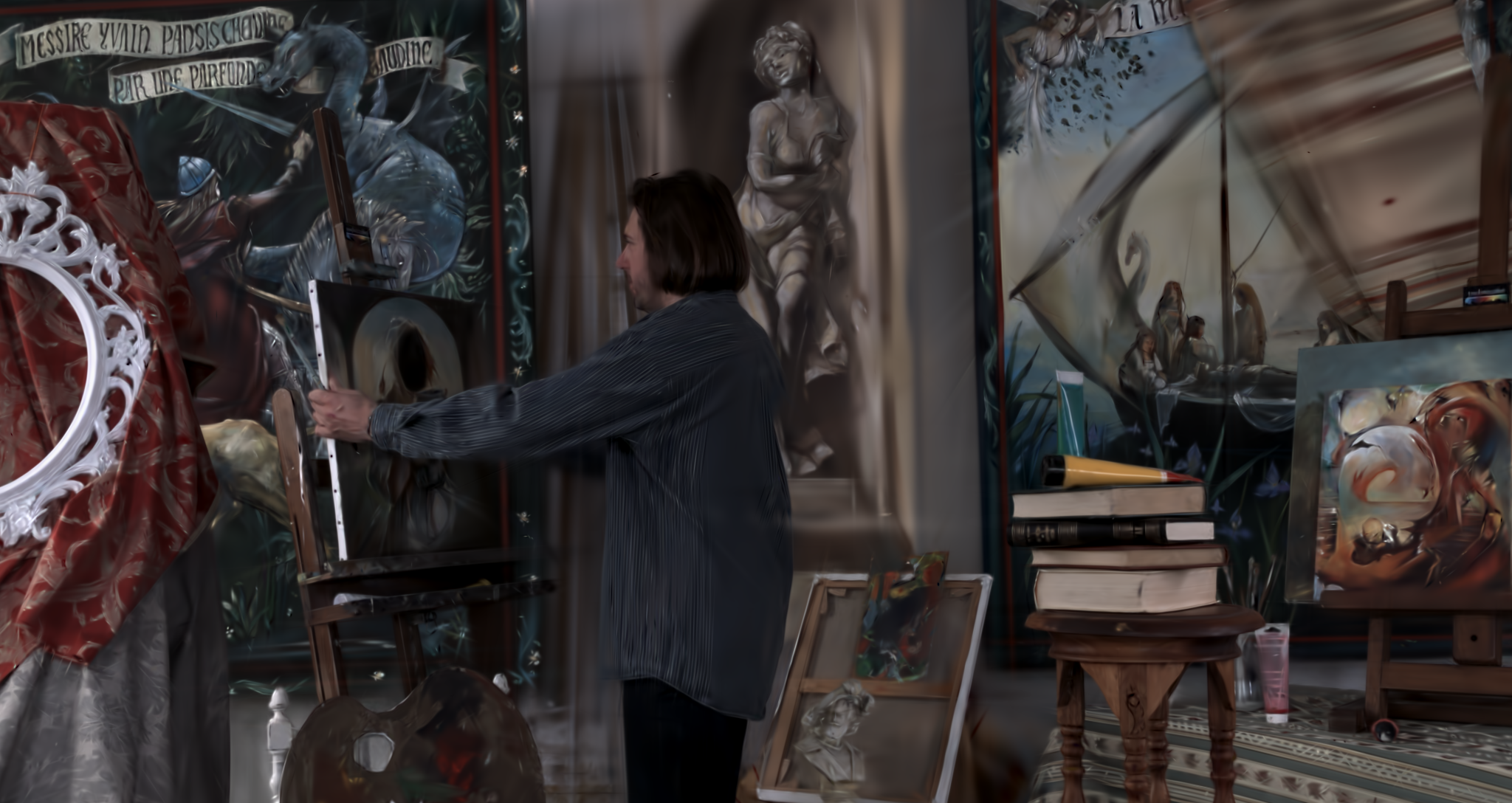}
        \includegraphics[width=\linewidth]{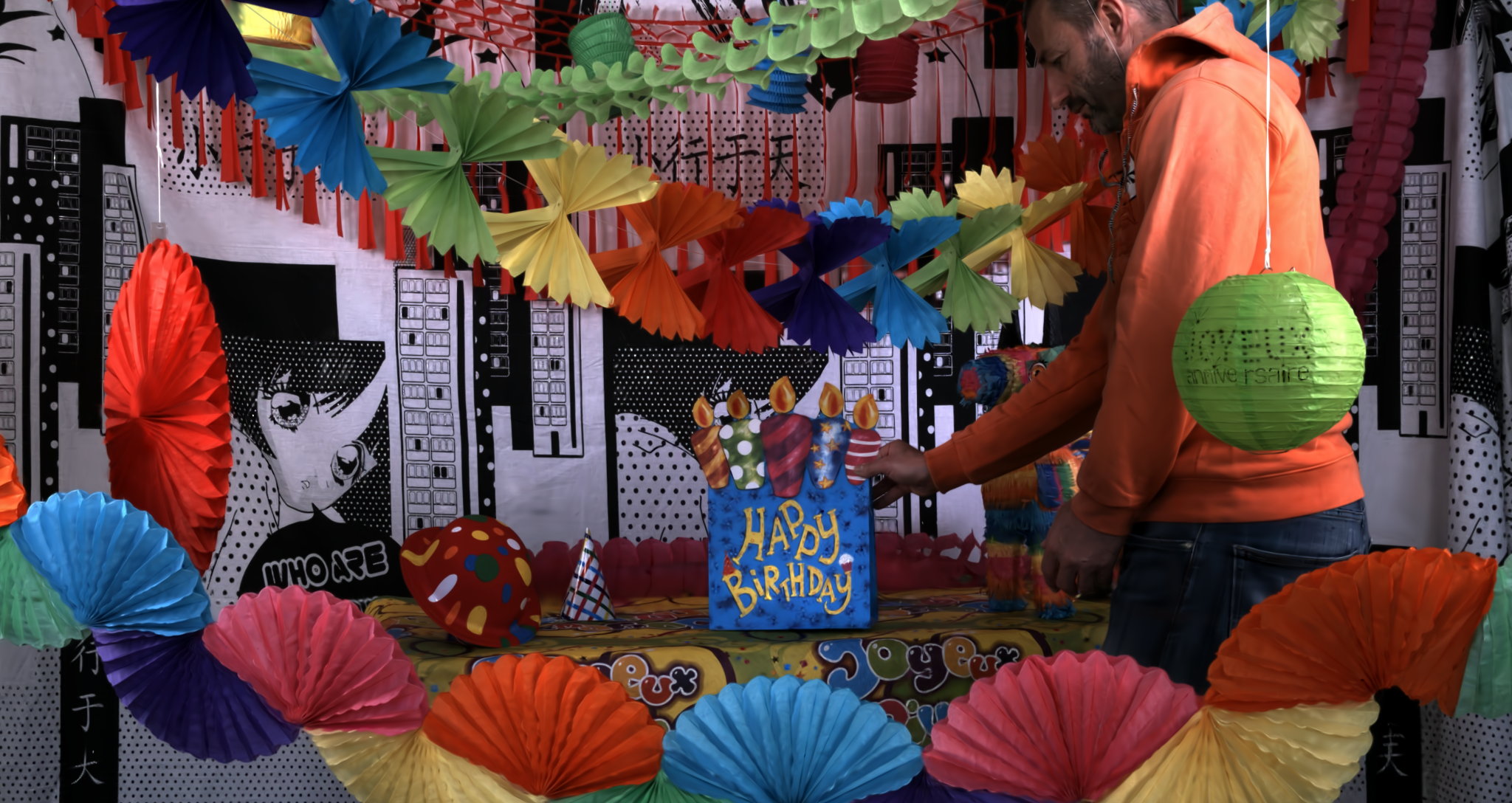}
        \includegraphics[width=\linewidth]{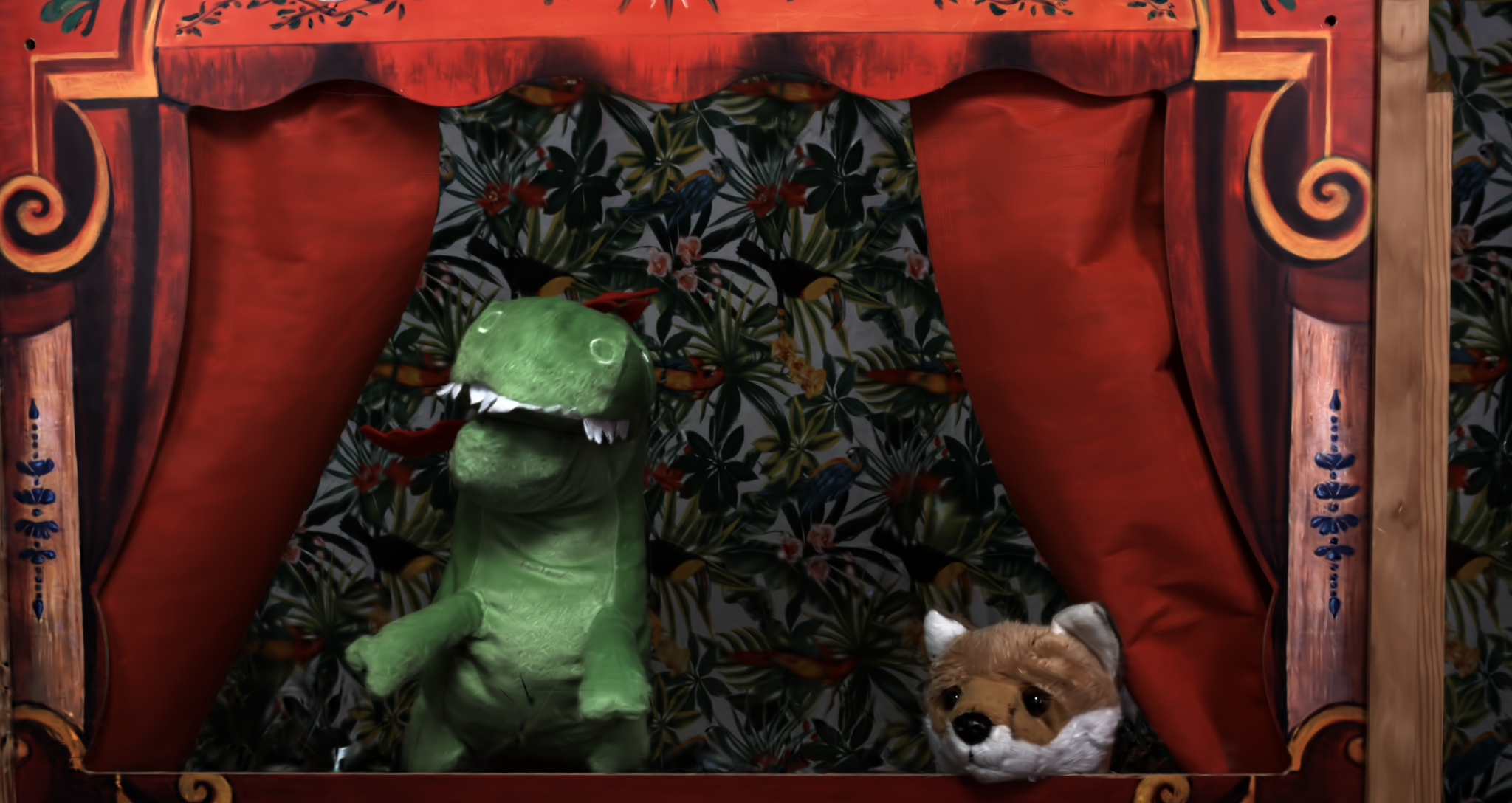}
        \includegraphics[width=\linewidth]{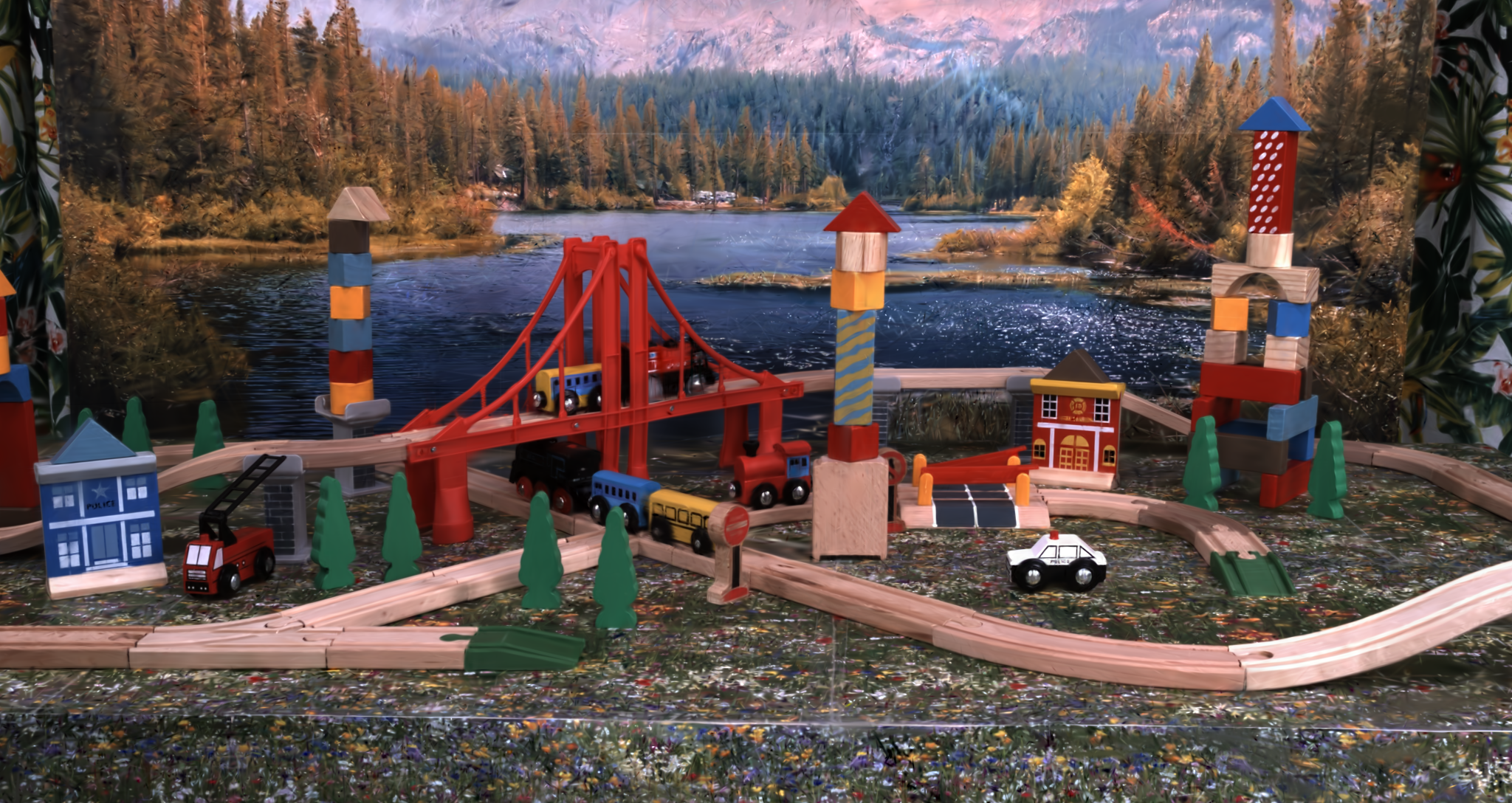}
        \subcaption{TC3DGS Render\\}
    \end{minipage}
    \vspace{0.3cm}
    \caption{\textbf{Qualitative Evaluation on the Technicolor dataset.} This figure shows a comparison between the ground truth and our reconstruction, demonstrating that our method achieves near-identical results despite extreme compression. }
    \label{fig:technicolor_visualizations}
\end{figure*}

To further demonstrate the effectiveness of our method, we evaluate our method on the Technicolor Dataset. The results in Table~\ref{tab:technicolor_perscene} show that our method provides competitive performance with a significantly smaller storage footprint. \Cref{fig:technicolor_visualizations} shows a comparison between the ground truth and our reconstruction, demonstrating that our method achieves near identical results.

\begin{table}[h!]
\caption{
\textbf{Per-Scene Results on the Technicolor Dataset.}
Our method achieves performance comparable to Dynamic3DGS \cite{luiten2023dynamic} while significantly reducing the model size. $\ast$STG needs multiple models for evaluation.}
\label{tab:technicolor_perscene}
\vspace{0.4cm}
\centering
\scalebox{0.9}{
\begin{tabular}{l|c|ccccc}
\toprule
Method & Avg. & \textit{Birthday} & \textit{Fabien} & \textit{Painter} & \textit{Theater} & \textit{Train} \\
\midrule
\midrule
\multicolumn{2}{l}{\textbf{PSNR$\uparrow$}} \\
\midrule
STG~\cite{stg} & $33.60$ & $32.09$ & $35.70$ & $36.44$ & $30.99$ & $32.58$ \\
D. 3DGS~\cite{luiten2023dynamic} & $32.12$ & $30.82$ & $33.62$ & $34.73$ & $31.12$ & $30.67$ \\
\midrule
 Ours &$31.9$ & $30.33$ & $33.15$ & $34.62$ & $31.01$ & $30.57$\\
\midrule
\midrule
\multicolumn{2}{l}{\textbf{SSIM$\uparrow$}} \\
\midrule
STG~\cite{stg} & - & - & - & - & - & - \\
D. 3DGS~\cite{luiten2023dynamic} & $0.88$ & $0.90$ & $0.88$ & $0.87$ & $0.86$ & $0.88$ \\
\midrule
Ours & $0.85$ & $0.88$ & $0.86$ & $0.81$	& $0.84$ & $0.86$ \\
\midrule
\midrule

\multicolumn{2}{l}{\textbf{LPIPS$\downarrow$}} \\
\midrule
STG~\cite{stg} & $0.08$ & $0.04$ & $0.11$ & $0.10$ & $0.13$ & $0.04$ \\
D. 3DGS~\cite{luiten2023dynamic} & $0.13$ & $0.07$ & $0.16$ &	$0.16$ & $0.14$ & $0.11$ \\
\midrule
Ours & $0.14$ & $0.09$ & $0.15$ &	$0.17$ & $0.16$ & $0.13$ \\
\bottomrule
\end{tabular}
}
\end{table}



        
        

\vspace{-0.1cm}
\section{Limitations}

Our approach has certain limitations rooted in the nature of dynamic 3D Gaussian splatting (3DGS) and the constraints of compressing temporal data for dynamic scenes. First, unlike compression approaches for static 3DGS, we are limited in our ability to aggressively compress position parameters, as doing so would compromise temporal consistency and spatial accuracy in the dynamic setting. Additionally, as our method builds on Dynamic 3DGS, it inherits its inability to accurately reconstruct new objects that enter the scene after the initial frame. This restricts its use in scenarios requiring complete adaptability to scene changes. Nonetheless, our method excels at handling scenes with complex and fast motions, maintaining high fidelity and rendering efficiency despite these constraints.

\begin{figure*}[t]

    \centering
    
    \hfill
    \begin{minipage}{0.49\linewidth}
        \centering
        \includegraphics[width=\linewidth]{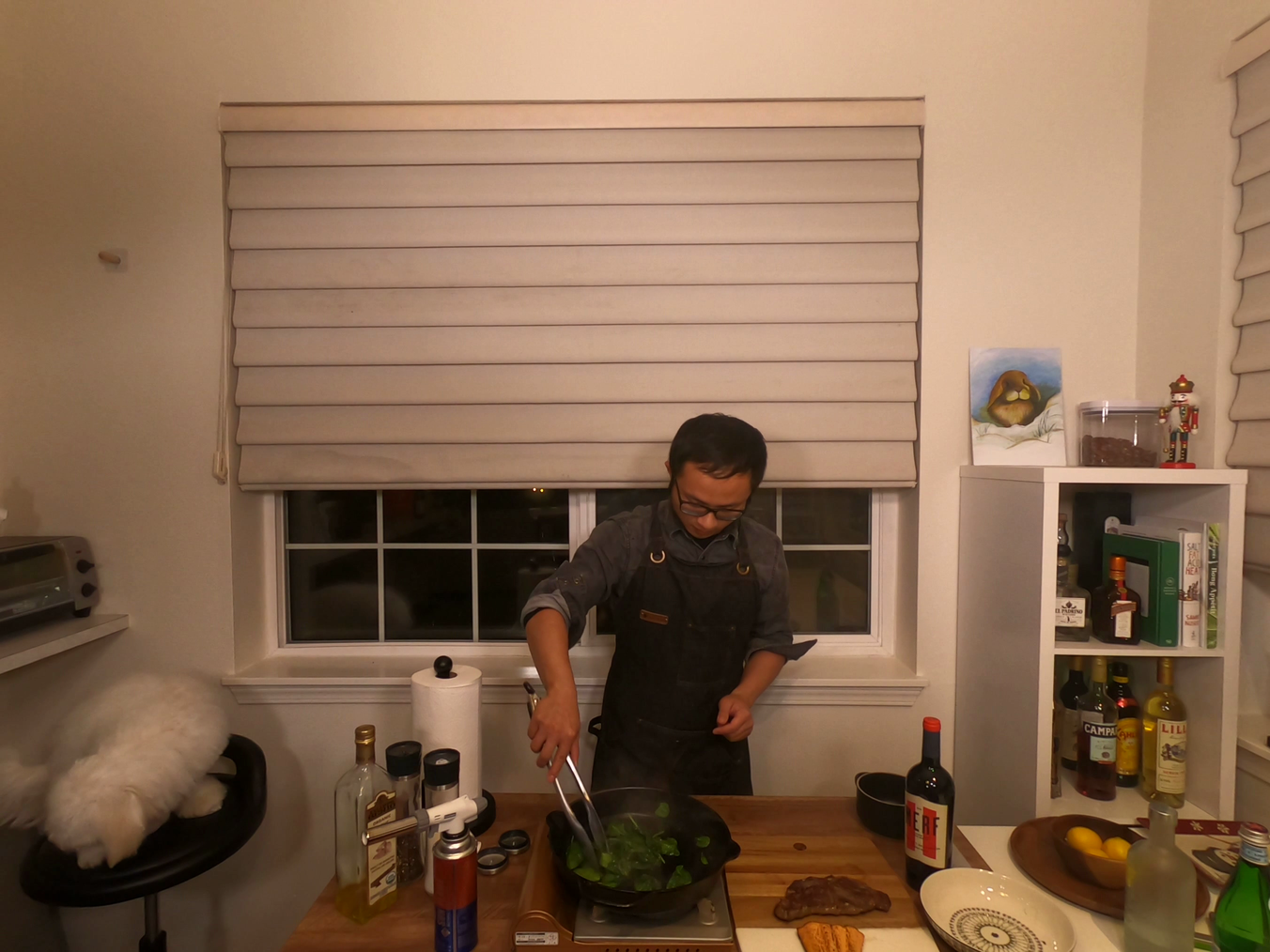}\vspace{0.2em}
        \includegraphics[width=\linewidth]{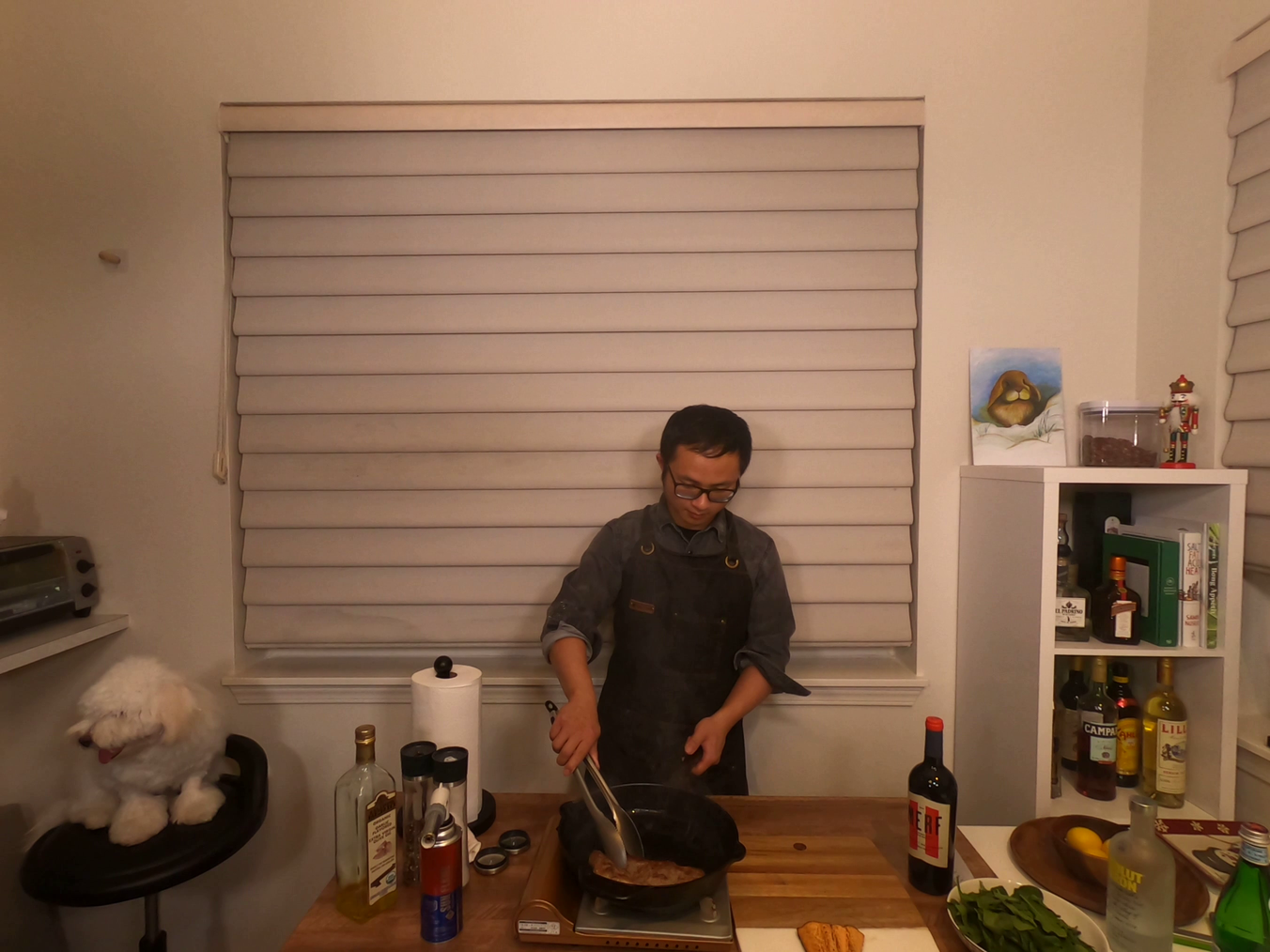}
        \subcaption{Ground-Truth}
    \end{minipage}
    \begin{minipage}{0.49\linewidth}
        \centering
        \includegraphics[width=\linewidth]{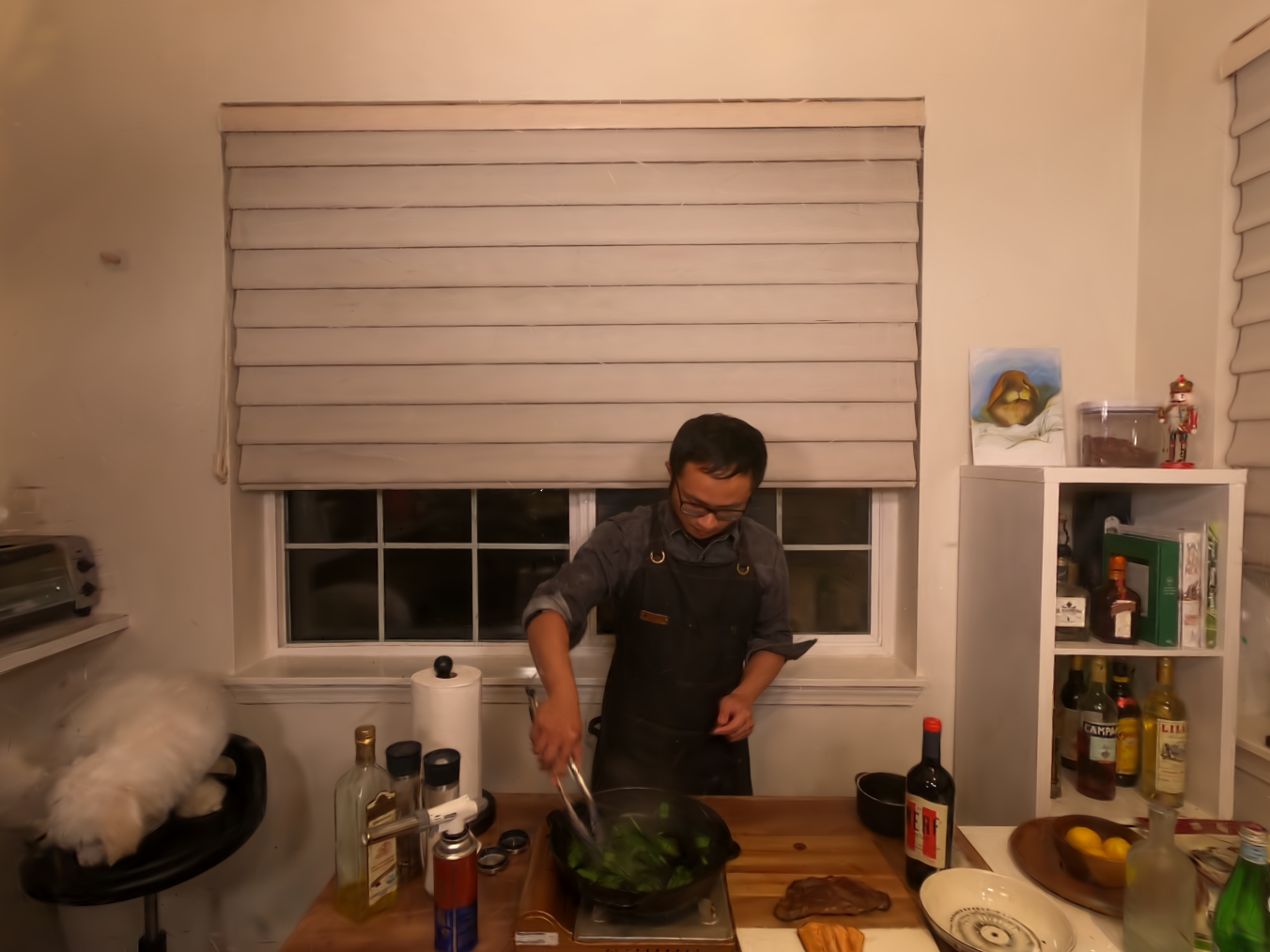}\vspace{0.2em}
        \includegraphics[width=\linewidth]{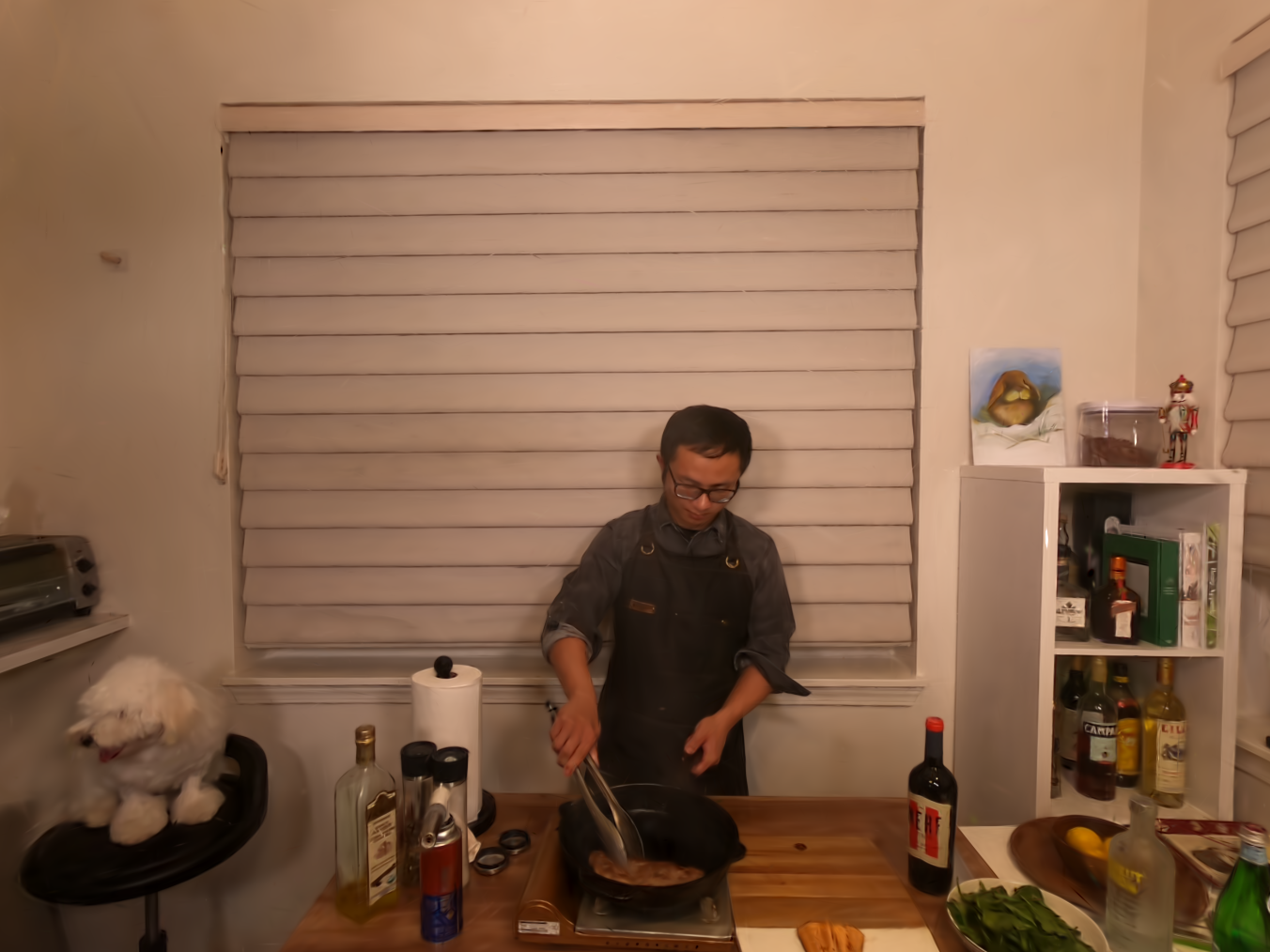}
        \subcaption{TC3DGS Render\\}
    \end{minipage}
    \vspace{0.3cm}
    \caption{\textbf{Qualitative Evaluation on the Neural 3D Video dataset.} This figure shows a comparison between the ground truth and our reconstruction, demonstrating that our method achieves near-identical results despite extreme compression. Top: Cook Spinach, Bottom: Sear Steak. }
    \label{fig:neural_3d_visualizations}
\end{figure*}

\vspace{-0.05cm}
\subsection{Datasets}
\label{sec:Datasets}
\vspace{-0.05cm}
\subsubsection{Panoptic Sports Dataset}
\vspace{-0.05cm}
We evaluate our method on the Panoptic Sports dataset, 
a subset of the Panoptic Studio dataset \cite{Joo_2019}. This dataset contains 6 different scenes, each having 31 camera sequences spanning 150 frames. We use 4 cameras (0, 10, 15 and 30) for testing and the rest for training, following the convention set by \cite{luiten2023dynamic}. In addition to the images, we use the provided foreground/background masks to apply a segmentation loss to improve the results and prevent the background from moving. This dataset contains complex motions with objects moving quickly and over long distances.

\subsubsection{Neural 3D Video Dataset}
The Neural 3D Video dataset \cite{li2022neural3dvideosynthesis} consists of 6 scenes with a number of cameras ranging from 18 to 21 and sequences of 300 frames. The images in this dataset are or resolution 2704$\times$2028, but we downsample them to 1352$\times$1014 for fair comparison with other methods. We hold out camera 0 for testing and do not apply any background loss, as background segmentations are not available for this dataset. 
\vspace{-0.1cm}

\clearpage

\end{document}